\PassOptionsToPackage{unicode}{hyperref}
\PassOptionsToPackage{hyphens}{url}
\documentclass[
  11pt,
]{article}
\usepackage{amsmath,amssymb}
\usepackage{iftex}
\ifPDFTeX
  \usepackage[T1]{fontenc}
  \usepackage[utf8]{inputenc}
  \usepackage{textcomp} 
\else 
  \usepackage{unicode-math} 
  \defaultfontfeatures{Scale=MatchLowercase}
  \defaultfontfeatures[\rmfamily]{Ligatures=TeX,Scale=1}
\fi
\usepackage{lmodern}
\ifPDFTeX\else
\fi
\IfFileExists{upquote.sty}{\usepackage{upquote}}{}
\IfFileExists{microtype.sty}{
  \usepackage[]{microtype}
  \UseMicrotypeSet[protrusion]{basicmath} 
}{}
\makeatletter
\@ifundefined{KOMAClassName}{
  \IfFileExists{parskip.sty}{%
    \usepackage{parskip}
  }{
    \setlength{\parindent}{0pt}
    \setlength{\parskip}{6pt plus 2pt minus 1pt}}
}{
  \KOMAoptions{parskip=half}}
\makeatother
\usepackage{xcolor}
\usepackage[margin=1in]{geometry}
\usepackage{longtable,booktabs,array}
\usepackage{calc} 
\usepackage{etoolbox}
\makeatletter
\patchcmd\longtable{\par}{\if@noskipsec\mbox{}\fi\par}{}{}
\makeatother
\IfFileExists{footnotehyper.sty}{\usepackage{footnotehyper}}{\usepackage{footnote}}
\makesavenoteenv{longtable}
\usepackage{graphicx}
\makeatletter
\def\maxwidth{\ifdim\Gin@nat@width>\linewidth\linewidth\else\Gin@nat@width\fi}
\def\maxheight{\ifdim\Gin@nat@height>\textheight\textheight\else\Gin@nat@height\fi}
\makeatother
\setkeys{Gin}{width=\maxwidth,height=\maxheight,keepaspectratio}
\makeatletter
\def\fps@figure{htbp}
\makeatother
\providecommand{\tightlist}{%
  \setlength{\itemsep}{0pt}\setlength{\parskip}{0pt}}
\ifLuaTeX
  \usepackage{selnolig}  
\fi
\usepackage{bookmark}
\IfFileExists{xurl.sty}{\usepackage{xurl}}{} 
\hypersetup{
  pdftitle={Mechanistic Analysis of Unified Proximity--Tactile Observation by a Capacitive Fingertip Array Constrained by a Metal Phalanx},
  pdfauthor={Shun Cheng},
  hidelinks,
  pdfcreator={LaTeX via pandoc}}

\title{Mechanistic Analysis of Unified Proximity--Tactile Observation by
a Capacitive Fingertip Array Constrained by a Metal Phalanx}
\author{Shun Cheng\thanks{Corresponding author. School of Computer Science and Software Engineering, Zhaoqing University, Zhaoqing, Guangdong, China. E-mail: \texttt{chengshun@zqu.edu.cn}}\\
School of Computer Science and Software Engineering,\\
Zhaoqing University,\\
Zhaoqing, Guangdong, China\\
\texttt{chengshun@zqu.edu.cn}}
\date{}

\begin{document}
\graphicspath{{figs/}{./}}

\maketitle

\textbf{Abstract---}Capacitive electrodes sense centimetre-scale conductive targets through fringing fields and contact force through dielectric compression, making them attractive as a shared proximity--tactile front-end on robotic fingertips. Mounted on a metal phalanx under a thin silicone skin, however, the electrode-to-body parasitic capacitance typically exceeds the useful proximity increment by one to three orders of magnitude, and crosstalk and time-division-multiplexing (TDM) settling prevent the capacitance-to-digital converter (CDC) reading from being identified with the electrode--target capacitance. This paper models the fingertip array as a multi-conductor system --- sensing electrodes, driven-shield nodes, the phalanx ground and an external target --- and derives the CDC measurand from the Maxwell capacitance matrix, with the termination of every inactive electrode written explicitly into the observation operator. Three coupled mechanisms are examined: (i) how a driven shield renders the shield-branch displacement current invisible to the CDC and how gain/phase mismatch leaks back as an equivalent capacitance; (ii) why far-field fringing and near-field compression require distinct observation operators; (iii) how the TDM settling time grows with the shield load and becomes a frame-rate constraint as the array scales. We further show that a single electrode layer used for both proximity and force reads \(\Delta C=\Delta C_{\mathrm{force}}+\Delta C_{\mathrm{dielectric}}(\varepsilon_{\mathrm{target}})\), so its force map cannot be material-independent, and analyse a dual-array stack in which the outer array is grounded as a Faraday screen during contact while the inner array reads the compression of the silicone gap. Electrostatic finite-element (FEM) simulations of this stack give a monotonic, strongly nonlinear contact response (4.6 fF at 0.05 mm to 370 fF at 0.5 mm compression on a 1.08 pF base), a nylon-versus-metal probe difference of at most 0.3 fF at the print resolution of the exported matrix (not established as mesh-converged), cancellation of the mutual-capacitance terms under driven-shield readout, and an electrical-crosstalk figure of about 1 fF at the convergence level of the current coarse mesh. In proximity mode the extracted electrode--shield coupling of 1.47 pF turns the \(10^{-3}\)--\(10^{-4}\) mismatch requirement into a quantitative gate: a 15.9 fF approach increment at 5 mm already equals the leakage at \(|\varepsilon_{A}|\approx10^{-2}\). The force map, mechanical crosstalk and all bench quantities remain unvalidated; the observation equations, FEM evidence and order-of-magnitude constraints form a verifiable baseline for prototype calibration, not measured performance claims.


\subsection{I. Introduction}\label{i.-introduction}

Dexterous manipulation requires two kinds of information that are contiguous in time and overlapping in space: the bearing and approach speed of the target before contact, and the normal force and pressure distribution after contact. Conventional designs assign proximity sensing to infrared, time-of-flight (ToF) or vision, and force sensing to piezoresistive, piezoelectric or strain-based structures. Hanging several external sensors on a finger pad only a few millimetres thick increases volume, wiring, calibration and power burdens at the same time. The attraction of capacitive structures is that one set of electrodes spans two physical regimes: when the electrode is much smaller than the target distance, the field is dominated by the fringing field and the reading rises as a conductive target approaches; when the electrode is separated by a thin dielectric layer and compressed, the field is approximately parallel-plate and the reading rises as the thickness decreases.

This dual capability is not new. NASA's Capaciflector already used an in-phase driven shield to push the field lines that would otherwise terminate on the robot body out into free space, for proximity sensing of manipulators {[}1{]}, {[}2{]}. Cheng et al.~cross-validated, by the method of images and finite elements, how the electrode-to-ground capacitance varies with distance, targeting metre-scale capacitive proximity fuzes {[}3{]}--{[}5{]}. In robotics, capacitive skins were used early on humanoid fingers and for large-area coverage: the iCub fingertips and palm employ capacitive pixels on flexible circuitry to sense contact pressure {[}6{]}, {[}7{]}; Lee et al.~switched between tactile and proximity modes on a single polymer platform by reconfiguring the electrode connections {[}8{]}. Proximity sensing has since been organised systematically into two use cases --- arm-level safety/interaction and in-hand pre-grasp/exploration {[}9{]} --- and capacitive proximity skins have been applied to gripper pre-alignment, liquid-state estimation and collaborative safety {[}10{]}--{[}12{]}. Recent work further integrates proximity and tactile sensing into the same flexible structure, e.g., a soft capacitive skin that resolves normal/shear components while detecting proximity at about 15 mm {[}13{]}, a fringing-field dual-mode flexible sensor {[}14{]}, and a three-mode array that trades resolution against detection depth with a switchable shield layer {[}15{]}. Fingertip-scale multi-axis force and distributed tactile sensing continue to develop in parallel {[}16{]}, {[}17{]}. Structurally closer to this paper are dual-layer self-capacitance schemes that switch between outer-electrode grounding for compression sensing and inner-electrode in-phase shielding for proximity {[}12{]}, {[}18{]}; bionic fingertips that ground the upper electrode after contact and time-multiplex the lower electrodes for self-capacitance force readout {[}19{]}; and capacitive interfaces that switch potentials between a measurement electrode and a guard electrode across an elastic dielectric to obtain approach and pressure in two steps {[}20{]}. These works show that dual-mode capacitive front-ends have already been demonstrated in hardware; Section X argues that an explicit observation operator, a quantitative mismatch gate and a TDM settling budget under a metal-phalanx constraint are still missing.

Despite this variety of paths, four mechanistic questions have not been isolated for miniaturised arrays mounted against a large-area metal phalanx.

First, a CDC never reads the ``electrode--target capacitance'' alone. What it reads is a combination of the Maxwell capacitance matrix of a multi-conductor system. Without shielding, the electrode--phalanx capacitance \(c_{iG}\) can reach tens to hundreds of times the useful proximity term \(\Delta C_{obj}\), and the relative sensitivity is drowned by the parasitic term. The role of a driven shield is not to delete the parasitic capacitance but to force the AC voltage difference across that branch to zero; once the follower exhibits a gain error \(\varepsilon_{A}\) or a phase lag, a leakage capacitance \(C_{leak}\) re-enters the measurement. Capacitive front-ends for collaborative robots already treat the active guard as a routine means of suppressing parasitics from tens of picofarads to nanofarads {[}11{]}, but few works state the mismatch constraint quantitatively against the useful signal \(\Delta C_{obj}\) and discuss it in a bone-mounted fingertip geometry.

Second, proximity and contact are not two ranges of the same observation operator. The far-field footprint is wider than the array pitch, so paralleling electrodes buys signal-to-noise ratio; the contact patch shrinks to one or two pixels and must be scanned point by point to form an image. Dynamic shielding or reconfigurable interconnection can switch between the two operators {[}8{]}, {[}15{]}, but the switching itself changes the boundary conditions of the neighbouring electrodes and hence the crosstalk matrix. If ``where the inactive electrodes are connected'' is not written into the observation model, array inversion is underdetermined.

Third, time-division multiplexing (TDM), adopted to reduce the number of op-amps and CDC channels, moves the error budget from space to time. The follower must drive the shield-to-ground capacitance \(C_{SG}\); insufficient settling appears as a channel gain that depends on the voltage of the previous time slot and is misread as proximity jitter or pixel crosstalk. The wiring and multiplexing constraints of large-area tactile skins have been reviewed specifically {[}21{]}; although a fingertip array is small, the same time constant quickly becomes the frame-rate ceiling at palm or arm scale.

Fourth, when the same electrode layer that looks outward for proximity is also used to measure force by compressing the skin above it, the contact reading contains the target's own permittivity. The force map then depends on what is being touched. This cross-sensitivity is structural: it cannot be calibrated away by a single material-independent force map, and it motivates separating the contact measurand from the open field by a grounded screen.

This paper gives no product specification, nor does it write unmeasured figures as conclusions. Its task is to build a multi-conductor observation model including the shield and the phalanx, and to condense the questions above into testable equations, order-of-magnitude constraints and FEM evidence. The contributions are:

\begin{itemize}
\tightlist
\item
  an explicit observation operator in which the termination of every inactive electrode (CIN, driven shield or ground) enters the CDC reading, including the driven-shield ``circulating'' readout under which mutual-capacitance terms cancel (Section II, eqs. (6a)--(6c));
\item
  a quantitative mismatch constraint for the driven shield, closed with FEM-extracted electrode--shield couplings for the proximity and contact operators (Section III-D);
\item
  a mechanistic account of the material cross-sensitivity of a single-layer proximity--force stack, and of its removal by a dual-array Faraday stack (L2 outer array grounded during contact, L4 inner array reading the L3 gap) (Sections II-E and VII-D);
\item
  a TDM implementation with explicit routing, atomic switching order and a locked settling window, analysed against the first-order settling model (Section VI);
\item
  electrostatic FEM evidence for the dual-array stack in all three operators, reported together with what it does not establish (Section VIII), and a numerical error budget (Section IX).
\end{itemize}

Section II gives the Maxwell matrix, the CDC reading and the dual-array operator; Section III analyses the driven shield and its mismatch; Section IV discusses far-field proximity and image-charge estimation; Section V gives the reconfigurable observation operators, the centroid and the crosstalk matrix; Section VI discusses TDM settling; Section VII analyses contact nonlinearity and material cross-sensitivity; Section VIII reports the FEM numerical validation; Section IX condenses noise and error terms into a budget; Section X discusses the relation to prior work and the implications for experiments. Dimensions appearing in the text --- 4 \(\times\) 4 elements, 2 mm pitch, skin thickness 0.8 mm --- are the FEM baseline geometry and serve as typical fingertip-scale values, not a frozen hardware bill of materials.

\subsection{II. Multi-Conductor Observation Model}\label{ii.-multi-conductor-observation-model}

\subsubsection{A. Range of Validity of the Parallel-Plate Formula}\label{a.-range-of-validity-of-the-parallel-plate-formula}

In the contact channel the electrode--dielectric spacing \(d\) is about 0.3--0.8 mm and the lateral extent is about 1.5--2 mm, so the field is still approximately uniform and the parallel-plate relation (1) holds; during contact the thickness reduction \(\Delta\)\(d\) dominates. In the proximity channel the other ``plate'' is an external target at a distance \(h\) of millimetres to tens of millimetres; \(h\) is much larger than the electrode size, the field is dominated by fringing, and (1) cannot be used to infer the distance. The CDC does not measure the single capacitance of (1) but the charge--voltage response of the whole conductor system. The multi-conductor model must therefore be written before shielding and inversion can be discussed.

\[C=\varepsilon_{0}\varepsilon_{r}\frac{A}{d},\qquad \frac{dC}{C}=\frac{d\varepsilon_{r}}{\varepsilon_{r}}+\frac{dA}{A}-\frac{dd}{d}\tag{1}\]

\subsubsection{B. Maxwell Capacitance Matrix}\label{b.-maxwell-capacitance-matrix}

We adopt the quasi-electrostatic approximation: the excitation frequency is 10 kHz--1 MHz and the module is much smaller than the wavelength. The conductor list contains \(N\) + 3 ideal conductors: the sensing electrodes \(E_{i}\) (\(i\) = 1, \ldots, \(N\)), the active shield \(S\), the metal phalanx (structural ground) \(G\), and the external target \(T\). The charge--voltage relation is

\[Q_{i}=\sum_{j} C_{ij}V_{j}.\tag{2}\]

The Maxwell capacitance matrix \(C\) is symmetric and positive semi-definite, with positive diagonal entries, non-positive off-diagonal entries and non-negative row sums. With the engineering positive-value convention

\[c_{ii}=\sum_{j} C_{ij},\qquad c_{ij}=-C_{ij}\ (i\neq j),\tag{3}\]

the charge equation becomes

\[Q_{i}=c_{ii}V_{i}+\sum_{j\neq i} c_{ij}(V_{i}-V_{j}).\tag{4}\]

Here \(c_{ii}\) is the self-capacitance of electrode \(i\) to ground with all other conductors grounded, and \(c_{ij}\) is the mutual capacitance between electrodes \(i\) and \(j\). Where no ambiguity arises, \(c_{ij}\) and \(C_{ij}\) are used interchangeably below; whenever signs matter, the convention of (3) is restored. Fig. 1 shows the resulting lumped equivalent circuit seen by the CDC.

\begin{figure}
\centering
\includegraphics{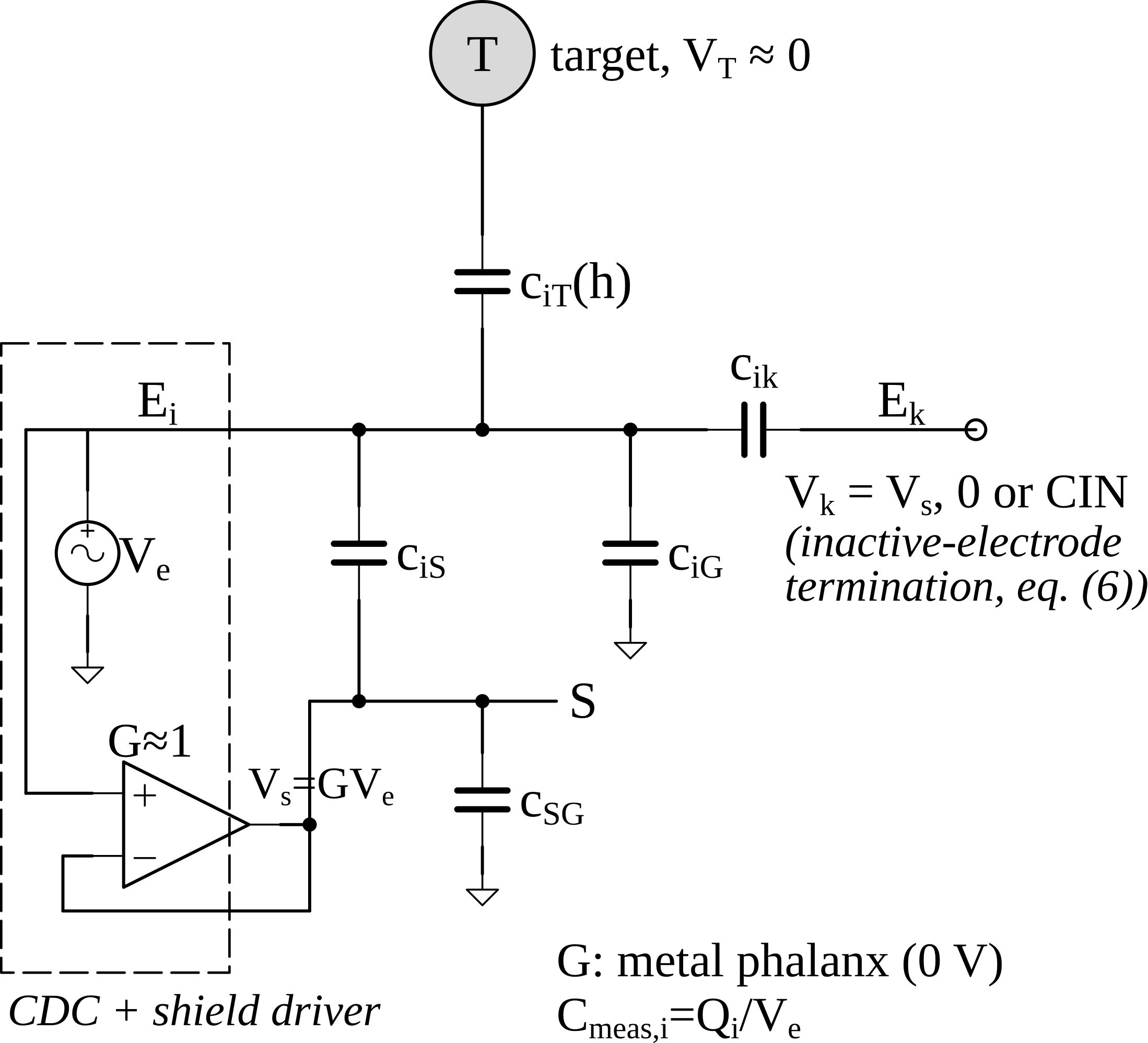}
\caption{Fig. 1}
\end{figure}

\emph{Fig. 1.} Multi-conductor equivalent circuit seen by the CDC for electrode \(E_{i}\). The CDC applies \(V_{e}\) and reads \(Q_{i}\); a unity-gain follower drives the shield node \(S\) to \(V_{s}=GV_{e}\) and supplies the shield-to-phalanx load \(c_{SG}\). The neighbour electrode \(E_{k}\) is terminated at \(V_{k}\in\{V_{s},0,\mathrm{CIN}\}\); this termination selects the instance of (6). Drawn in KiCad; component values are not implied.

\subsubsection{C. What the CDC Actually Measures}\label{c.-what-the-cdc-actually-measures}

In self-capacitance measurement an excitation \(V_{e}\) is applied to the electrode under test, the shield is tied to \(V_{s}\), and the phalanx is grounded (\(V_{G}\) = 0). With the definition

\[I_{i}=\frac{dQ_{i}}{dt},\qquad C_{\mathrm{meas},i}\equiv\frac{Q_{i}}{V_{e}}.\tag{5}\]

When the target can be approximated as an AC ground at the excitation frequency (\(V_{T}\) \(\approx\) 0, consistent with the treatment in {[}1{]}),

\[C_{\mathrm{meas},i}=C_{ii}+C_{iS}\frac{V_{s}}{V_{e}}+\sum_{k\neq i} C_{ik}\frac{V_{k}}{V_{e}}.\tag{6}\]

Equation (6) is the observation equation of this paper: ``measuring capacitance'' means measuring the sum of the self-capacitance term and all mutual terms weighted by the neighbouring-conductor potentials. Without shielding, with neighbouring electrodes grounded and \(V_{s}\) = 0, (6) degenerates to

\[C_{\mathrm{meas},i}=c_{iT}(h)+c_{iG}+c_{iS}+c_{\mathrm{wire}}+\sum_{k} c_{ik}.\tag{7}\]

The only useful term is \(c_{iT}\)(\(h\)). Mounted on a metal phalanx, \(c_{iG}\) can exceed \(c_{iT}\) by one to three orders of magnitude, and the relative sensitivity

\[S=\frac{\Delta C_{\mathrm{obj}}}{C_{\mathrm{para}}+C_{\mathrm{obj}}}\approx\frac{\Delta C_{\mathrm{obj}}}{C_{\mathrm{para}}}\quad(C_{\mathrm{para}}\gg C_{\mathrm{obj}})\tag{8}\]

is suppressed by the parasitic term. This is why an unshielded array ``sees the body but not the object'': the proximity field is not absent --- it rides on a much larger bias capacitance, and its relative change falls close to the CDC noise floor.

\subsubsection{D. Grounded Versus Floating Targets}\label{d.-grounded-versus-floating-targets}

For a grounded target (\(V_{T}\) = 0), \(c_{iT}\) enters (6) directly. For a floating target (\(Q_{T}\) = 0), eliminating \(V_{T}\) shows that the electrode sees the series combination of the electrode--target and target--ground capacitances; the typical form given for the Capaciflector is {[}1{]}

\[C_{1g}=C_{sg}+\frac{C_{so}C_{og}}{C_{so}+C_{og}}.\tag{9}\]

At the human-body scale \(C_{og}\) \(\gg\) \(C_{so}\), so the series term approaches \(C_{so}\), of the same order as for a grounded target; this explains the strong response of capacitive proximity sensing to the human body. For non-conductors such as dry plastics, \(C_{og}\) is small and unstable, and the proximity channel carries no theoretical guarantee comparable to that for conductive targets. The order-of-magnitude discussion below assumes conductive targets by default. A dielectric target nevertheless perturbs the field through its permittivity; Section VII-D shows that this is precisely what contaminates a force reading taken in the open field.

\subsubsection{E. Dual-Array Stack and the Termination-Dependent Observation Operator}\label{e.-dual-array-stack-and-the-termination-dependent-observation-operator}

The analysed stack (Fig. 2) has five functional layers above the phalanx: L1 silicone skin (0.8 mm, \(\varepsilon_{r}\) = 3.0), L2 outer electrode array, L3 compressible silicone gap (\(d_{0}\) = 0.6 mm, \(\varepsilon_{r}\) = 2.7), L4 inner electrode array, and L5 continuous bottom shield; L6 is a process layer (stiffener/coverlay) without an electrical role. L2 and L4 carry identical, concentric 4 \(\times\) 4 arrays (pad 1.4 mm, pitch 2.0 mm; Fig. 3). Neither array has a fixed function. Each of the \(2N\) electrodes is routed, slot by slot, to one of four buses --- the CDC inputs CIN\_A or CIN\_B, the driven shield SHLD, or GND --- so that ``shield'' is a role rather than a dedicated conductor.

\begin{figure}
\centering
\includegraphics{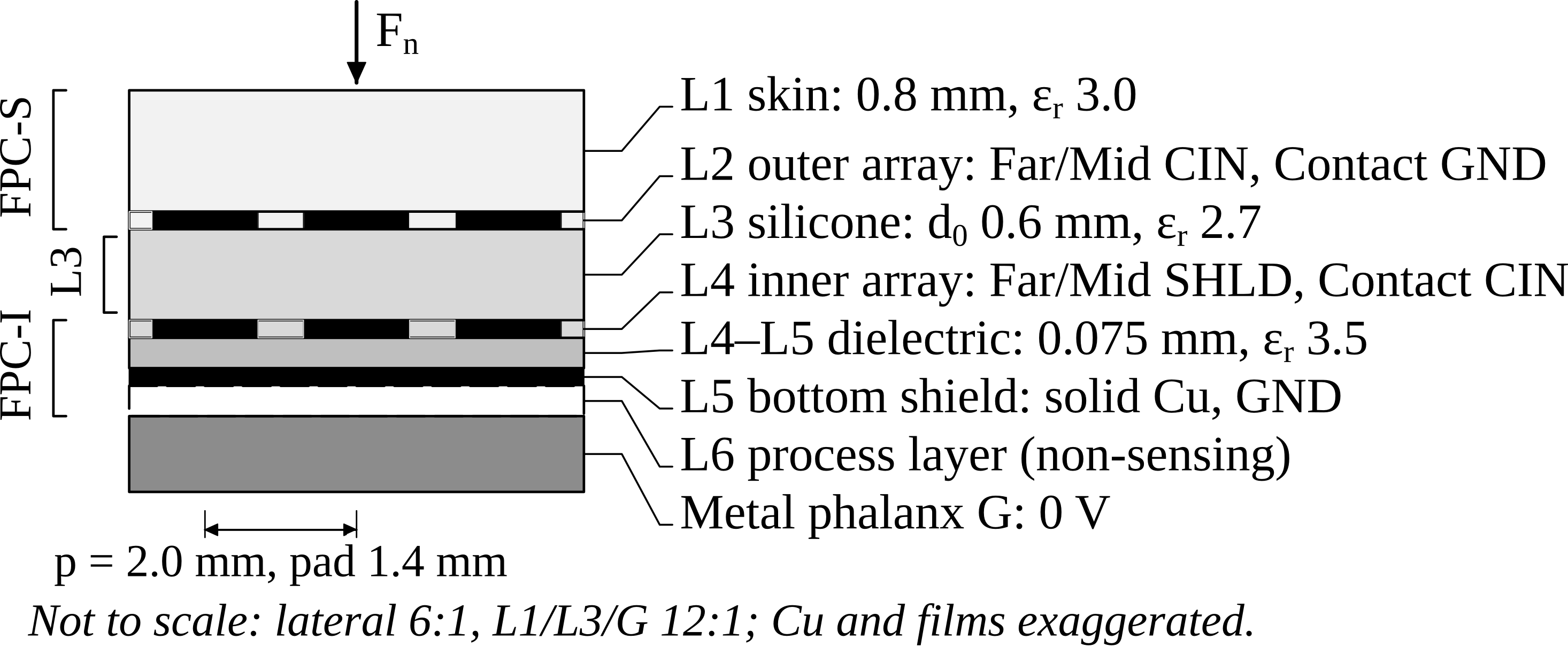}
\caption{Fig. 2}
\end{figure}

\emph{Fig. 2.} Cross-section of the dual-array stack (FEM baseline geometry, not to scale). L2 and L4 change role with the operating mode; L5 is a continuous ground (or shield) pour; L6 is a non-sensing process layer. FPC-S and FPC-I indicate the two flexible circuits of the engineering package; L3 is laminated manually between them. Drawn in KiCad.

\begin{figure}
\centering
\includegraphics{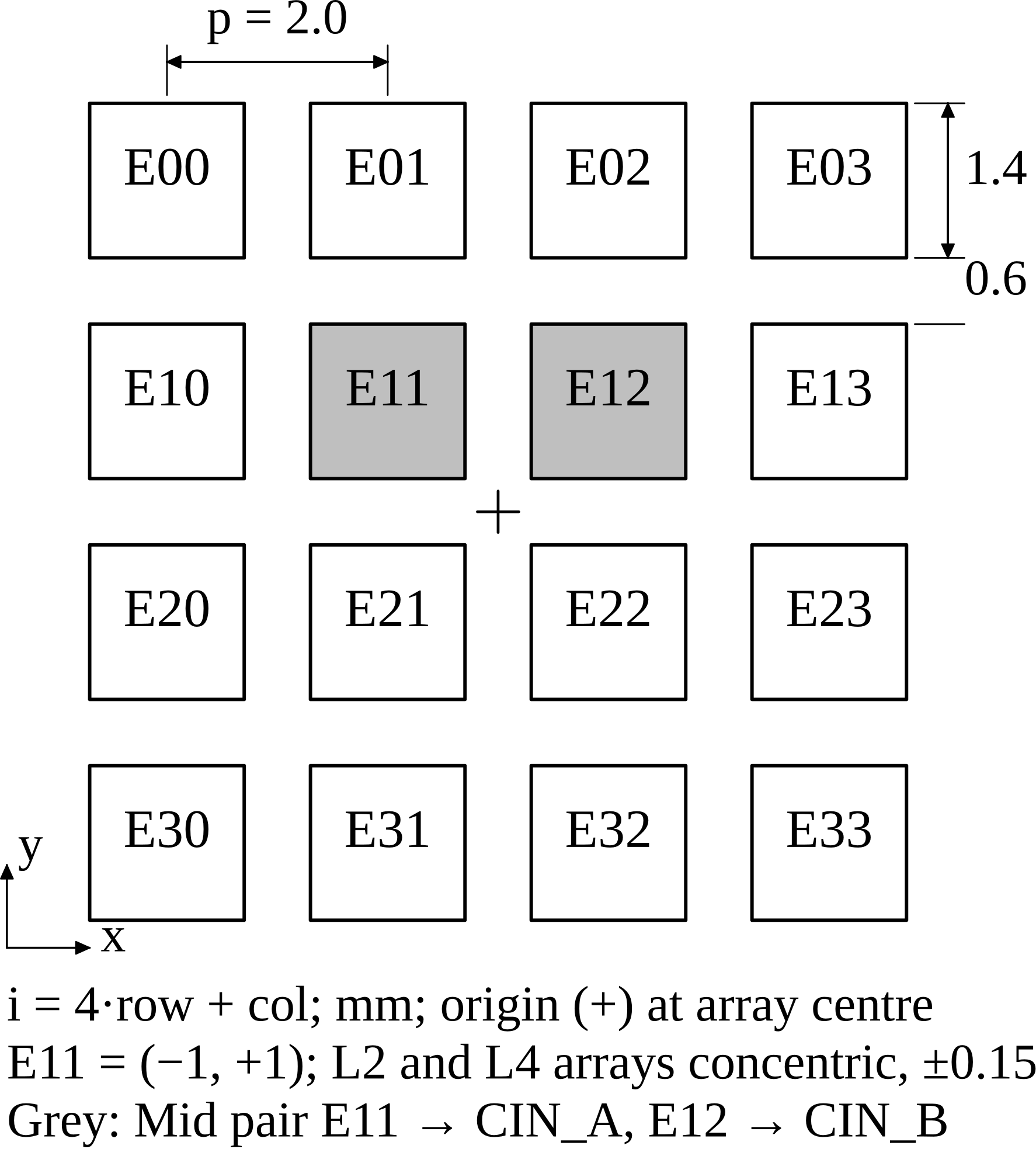}
\caption{Fig. 3}
\end{figure}

\emph{Fig. 3.} Electrode layout (identical on L2 and L4). Electrode \(E_{rc}\) has index \(i=4r+c\); \(E_{11}\) lies at \((-1,+1)\) mm from the array centre. The grey pair \(E_{11}/E_{12}\) is the default Mid hardware-differential pair. Drawn in KiCad.

Let \(\mathcal{A}\) be the set of electrodes connected to the CDC input under test, \(\mathcal{S}\) the set tied to the shield bus, and \(\mathcal{Z}\) the set tied to GND together with L5, the phalanx and the target. Summing (4) over \(\mathcal{A}\) with \(V_{k}=V_{e}\) for \(k\in\mathcal{A}\), \(V_{k}=V_{s}\) for \(k\in\mathcal{S}\) and \(V_{k}=0\) for \(k\in\mathcal{Z}\) gives the termination-dependent form of (6):

\[C_{\mathrm{meas},\mathcal{A}}=\sum_{i\in\mathcal{A}}\Big[\sum_{k\in\mathcal{Z}}c_{ik}+\Big(1-\frac{V_{s}}{V_{e}}\Big)\sum_{k\in\mathcal{S}}c_{ik}\Big].\tag{6a}\]

With an ideal follower (\(V_{s}=V_{e}\)) the second bracket vanishes: the reading is the direct capacitance from the active set to every grounded conductor, and every coupling between the active set and shield-tied electrodes --- including the mutual capacitance between neighbouring pads --- drops out. We call this the circulating (``circ'') readout. In terms of the Maxwell entries exported by a field solver it is the row sum over all conductors held at \(V_{e}\),

\[C_{\mathrm{circ},\mathcal{A}}=\sum_{i\in\mathcal{A}}\Big(C_{ii}+\sum_{k\in\mathcal{A}\cup\mathcal{S},\,k\neq i}C_{ik}\Big),\tag{6b}\]

which differs from the raw diagonal entry \(C_{ii}\) by the (negative) shield couplings. The distinction is not cosmetic: Section VIII-C shows that a differential channel evaluated with raw diagonals loses most of its signal, and that the circ form recovers it.

The three operating modes are three choices of \((\mathcal{A},\mathcal{S},\mathcal{Z})\) (Table I, Fig. 4). In Contact, all of L2 is placed in \(\mathcal{Z}\). L2 then forms a grounded screen between the active inner pad and everything above the skin, so that (6a) for an active L4 pad \(i\) becomes

\[C_{\mathrm{meas},i}^{\mathrm{Contact}}=c_{i,\mathrm{L2}}(d_{0}-\Delta d)+c_{i,\mathrm{L5/G}}+c_{iT}^{\mathrm{res}},\qquad c_{iT}^{\mathrm{res}}\to0\ \text{for a closed L2 screen}.\tag{6c}\]

The first term carries the L3 compression; the target appears only through the residual \(c_{iT}^{\mathrm{res}}\) that leaks through gaps between the L2 pads and, in hardware, through seams and the finite impedance of the L2 ground return. Equation (6c) is the Faraday observation operator. Its premise --- that the residual is small --- is a geometric claim tested by FEM in Section VIII-D, under the idealisation of an equipotential L2 sheet.

\textbf{TABLE I} Node Roles per Operating Mode (V2 Routing)

\begin{longtable}[]{@{}
  >{\raggedright\arraybackslash}p{(\columnwidth - 8\tabcolsep) * \real{0.2000}}
  >{\raggedright\arraybackslash}p{(\columnwidth - 8\tabcolsep) * \real{0.2000}}
  >{\raggedright\arraybackslash}p{(\columnwidth - 8\tabcolsep) * \real{0.2000}}
  >{\raggedright\arraybackslash}p{(\columnwidth - 8\tabcolsep) * \real{0.2000}}
  >{\raggedright\arraybackslash}p{(\columnwidth - 8\tabcolsep) * \real{0.2000}}@{}}
\toprule\noalign{}
\begin{minipage}[b]{\linewidth}\raggedright
Mode
\end{minipage} & \begin{minipage}[b]{\linewidth}\raggedright
L2 (outer)
\end{minipage} & \begin{minipage}[b]{\linewidth}\raggedright
L4 (inner)
\end{minipage} & \begin{minipage}[b]{\linewidth}\raggedright
CDC measurand
\end{minipage} & \begin{minipage}[b]{\linewidth}\raggedright
Use
\end{minipage} \\
\midrule\noalign{}
\endhead
\bottomrule\noalign{}
\endlastfoot
Far & all 16 \(\rightarrow\) CIN\_B (paralleled) & all \(\rightarrow\) SHLD & self-capacitance at CIN\_B, (6a) & presence, coarse distance, \textbar d\(C\)/d\(t\)\textbar{} \\
Mid & \(E_{11}\) \(\rightarrow\) CIN\_A, \(E_{12}\) \(\rightarrow\) CIN\_B, others \(\rightarrow\) SHLD & all \(\rightarrow\) SHLD & hardware differential \(C_{A}-C_{B}\) & lateral centring \\
Contact & all 16 \(\rightarrow\) GND (Faraday screen) & \(L4[k]\) \(\rightarrow\) CIN\_A, other 15 \(\rightarrow\) SHLD & self-capacitance at CIN\_A, (6c) & pressure image, \(F_{n}\) \\
\end{longtable}

\begin{figure}
\centering
\includegraphics{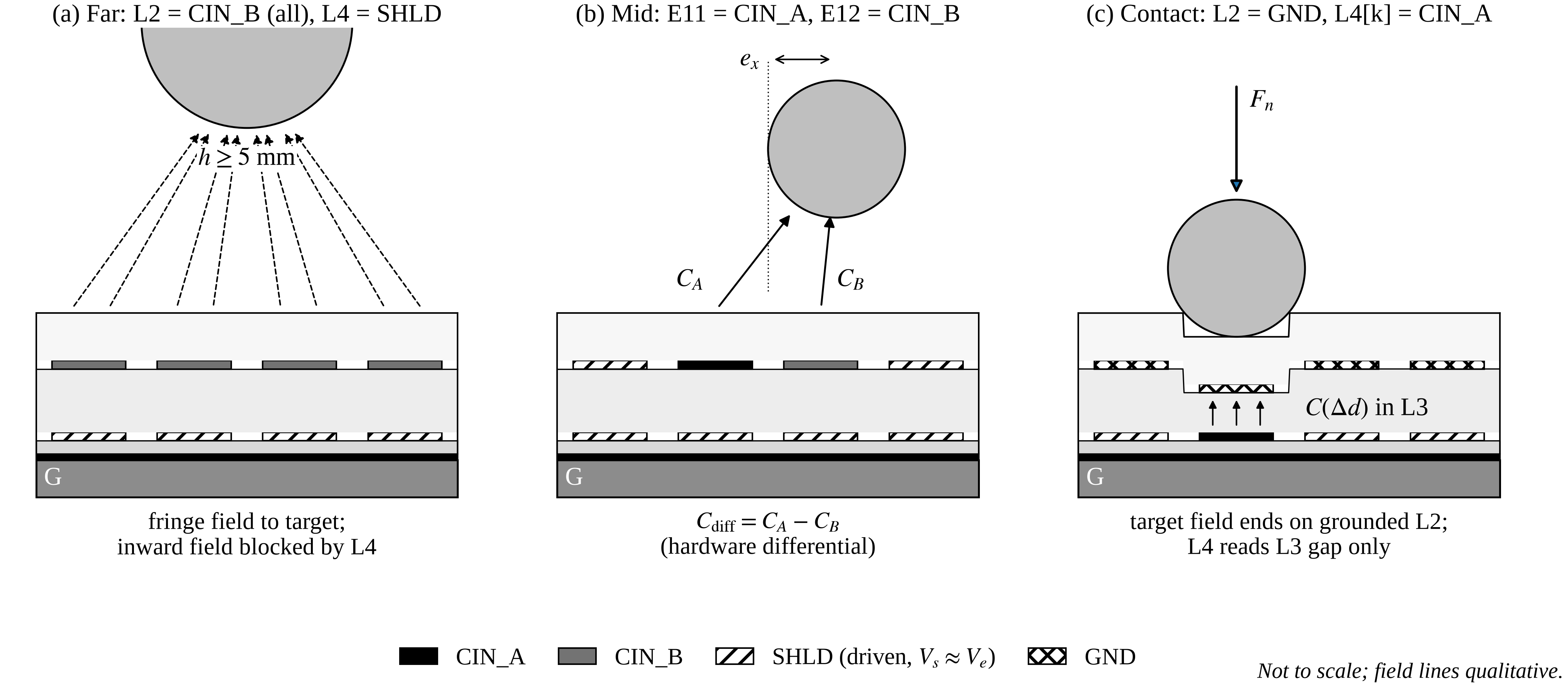}
\caption{Fig. 4}
\end{figure}

\emph{Fig. 4.} The three observation operators of the dual-array stack, drawn as a section through row 1 (\(E_{10}\)--\(E_{13}\)). (a) Far: paralleled L2 pads look outward while L4 blocks the inward field. (b) Mid: hardware differential between \(E_{11}\) and \(E_{12}\). (c) Contact: L2 is grounded; the target field ends on L2 and L4{[}\(k\){]} reads only the L3 gap. Field lines are qualitative (matplotlib schematic).

\subsection{III. Driven Shield and Mismatch Mechanism}\label{iii.-driven-shield-and-mismatch-mechanism}

\subsubsection{A. Equipotential Condition}\label{a.-equipotential-condition}

A driven shield does not remove \(c_{iG}\) from the stack-up; it forces the AC voltage difference across that branch to zero, makes its displacement current invisible to the CDC, and pushes the electric field out of the body. Electrostatically there are three layers: the active electrode at \(V_{e}\), the remaining pixels and the intermediate shield at \(V_{s}\), and the phalanx at \(V_{G}\) = 0. The shield is driven by a voltage follower,

\[V_{s}(t)=G\,V_{e}(t),\qquad G\approx 1.\tag{10}\]

When \(V_{s}\) = \(V_{e}\), the shield current

\[I_{eS}=C_{eS}\frac{d}{dt}(V_{e}-V_{s})=0,\tag{11}\]

vanishes and is invisible to \(C_{meas,i}\) as defined in (5). No net field exists between equipotential surfaces: the field lines pointing toward the phalanx are truncated by the shield and can only escape outward from the electrode's top surface and the gaps. With ideal shielding and neighbouring electrodes clamped at \(V_{s}\) or ground, the reading reduces to

\[C_{\mathrm{meas},i}\approx c_{iT}(h)+c_{i,\mathrm{fringe}}+c_{\mathrm{wire,unsh}}.\tag{12}\]

The fringing term is set by the shield's outward extension. Too small an extension lets side field lines bend back toward the phalanx; too large an extension occupies finger-pad area and increases \(C_{SG}\). For a fingertip array with a pitch of about 2 mm, an extension on the order of one pitch is a compromise between geometry and load, not a process dogma that must be obeyed. From the field-partitioning viewpoint, the field lines in the electrode--shield interlayer cancel; those in the electrode--target region are pushed outward; and those in the shield--phalanx region are dense, corresponding to a \(C_{SG}\) of up to tens of picofarads. The latter capacitance never enters the CDC signal path, but it is driven entirely by the follower and therefore enters the settling-time analysis of Section VI. In the dual-array stack the inner array L4 acts as this shield in Far and Mid modes; because L4 consists of pads rather than a solid plane, the gaps between L4 pads expose the grounded L5 pour, and that path remains in \(\mathcal{Z}\) of (6a).

\subsubsection{B. Gain and Phase Mismatch}\label{b.-gain-and-phase-mismatch}

A practical follower has a gain error \(\varepsilon_{A}\) and a phase lag \(\varphi\):

\[V_{s}=(1+\varepsilon_{A})e^{-j\varphi}V_{e}\ \Rightarrow\ V_{e}-V_{s}\approx-(\varepsilon_{A}+j\varphi)V_{e}.\tag{13}\]

The equivalent capacitance leaking into the measurement is

\[C_{\mathrm{leak}}\equiv\mathrm{Re}\!\left[\frac{I_{eS}}{j\omega V_{e}}\right]\approx C_{eS}\,\varepsilon_{A}.\tag{14}\]

In a fingertip stack-up \(C_{eS}\) is typically a few to tens of picofarads. If the useful Far-mode increment \(\Delta C_{obj}\) is only 10--200 fF, then

\[|\varepsilon_{A}|\sim 10^{-3}\text{--}10^{-4},\tag{15}\]

is required, otherwise the leakage term is comparable to the useful signal. Equation (15) is the first constraint on whether bone-mounted proximity sensing can work at all. It also shows that ``shorting the shield layer to the electrode at DC'' does not constitute a driven shield: a short guarantees a common potential only at low frequency --- it cannot guarantee \(V_{s}\) = \(V_{e}\) with phase synchrony at the excitation frequency, and it cannot drive \(C_{SG}\). A follower or the CDC's dedicated shield port must be used, and the follower must not slew-rate-limit at the excitation frequency. The phase term \(\varphi\) gives \(I_{eS}\) a component in quadrature with the excitation. A synchronous-demodulation CDC is insensitive to the quadrature component, but with a wideband front-end or a distorted excitation waveform the phase mismatch still appears as a frequency dependence of the apparent capacitance. Therefore \(\varepsilon_{A}\) and \(\varphi\) should be calibrated separately: the former corresponds to in-phase leakage; the latter to a reactive current that can be suppressed by demodulation but still consumes follower headroom. Comparing (14) with (6a) identifies \(C_{eS}\) with \(\sum_{i\in\mathcal{A}}\sum_{k\in\mathcal{S}}c_{ik}\), i.e.~with the magnitude of the electrode--shield entry of the Maxwell matrix; Section III-D uses this identity to evaluate (14) from FEM.

\subsubsection{C. Boundary Conditions of Inactive Electrodes}\label{c.-boundary-conditions-of-inactive-electrodes}

While electrode \(i\) is measured, the other electrodes have three possible terminations, corresponding to three different instances of (6). Grounded: the neighbours become additional grounds, their mutual terms enter the reading, and the proximity field is pulled toward the array interior. Floating: their potentials are set by leakage resistances and residual mutual capacitances, with poor repeatability. Tied to \(V_{s}\): the neighbours become lateral shields, their mutual terms are invisible to the CDC, but the effective aperture of the array moves with the measured pixel.

The last termination generalises the same potential logic as the driven shield: the current pixel acts as the sensor and all other pixels as part of the shield. It lets \(N\) electrodes share one follower and one CDC, at the cost of re-establishing \(V_{s}\) at every switching event, and the crosstalk matrix becomes a function of the switch state. The inversion of Section V must be consistent with this boundary condition: applying a crosstalk matrix calibrated in ``grounded-scan'' mode to an operating state with neighbours tied to \(V_{s}\) is a systematic bias, not random noise. The dual-array stack adds the fourth termination of Table I --- an entire layer grounded as a screen --- which is not a neighbour boundary condition but a change of which conductors can be seen at all.

Routing geometry obeys the same principle: signal conductors should fall within the shield's projection, otherwise the tail traces re-introduce picofarad-level capacitance to ground, and \(c_{wire,unsh}\) in (12) cancels the shielding gain. This is a constraint of the observation model on routing, not a stack-up process checklist.

\subsubsection{D. FEM-Extracted Shield Coupling and the Quantitative Mismatch Constraint}\label{d.-fem-extracted-shield-coupling-and-the-quantitative-mismatch-constraint}

The empty-domain baselines of the FEM campaign (Section VIII) give the electrode--shield entry directly. In the Far configuration (16 L2 pads paralleled on CIN\_B, all L4 on SHLD, no target) the Maxwell entries are \(C_{\mathrm{CIN_B}}\) = 2.6168 pF, \(C_{\mathrm{SHLD}}\) = 17.092 pF and \(C_{\mathrm{CIN_B,SHLD}}\) = -1.467 pF, so \(C_{eS}\) \(\approx\) 1.467 pF. This value is consistent with a parallel-plate estimate from (1) for 16 pads of 1.4 mm \(\times\) 1.4 mm across the 0.6 mm L3 gap (\(\varepsilon_{r}\) = 2.7), which gives about 1.25 pF before fringing. In the Contact configuration (L2 grounded, L4{[}\(E_{11}\){]} on CIN\_A, the other 15 L4 pads on SHLD) the entry is only \(C_{\mathrm{CIN_A,SHLD}}\) = -10.2 fF, because the shield-tied neighbours are coplanar pads rather than an overlapping layer; the sum of the individually resolved neighbour mutual capacitances in the 16-port contact model (Section VIII-E) is 10.1 fF at zero compression, in agreement.

Inserting these values into (14) gives the model prediction of Fig. 5. In Far mode, \(|\varepsilon_{A}|=10^{-3}\) leaks 1.47 fF, half of the 3.1 fF increment from \(h\) = 20 mm to 10 mm and about one tenth of the 15.9 fF increment from 20 mm to 5 mm (Section VIII-B). Holding the leakage below 10 \% of these increments requires \(|\varepsilon_{A}|\lesssim2\times10^{-4}\) and \(\lesssim1.1\times10^{-3}\), respectively --- inside the band of (15), now with the stack-specific \(C_{eS}\). In Contact mode the same \(|\varepsilon_{A}|=10^{-3}\) leaks about 0.01 fF, more than two orders of magnitude below the smallest simulated compression increment (4.6 fF). The mismatch constraint is therefore a Far-mode constraint in this stack; Contact is limited by other terms (Section IX). These are analytical predictions using FEM-extracted couplings; the follower gain error itself has not been measured.

A first-order estimate already questions whether the AD8606 follower assumed in Section VI-B can meet the Far gate. The FDC1004 excitation is fixed at 25 kHz {[}22{]}; the AD8606 unity-gain bandwidth is about 10 MHz. Approximating the closed-loop magnitude error of a unity-gain follower by

\[|\varepsilon_{A}|\sim \frac{f_{\mathrm{exc}}}{\mathrm{GBW}}=\frac{25\ \mathrm{kHz}}{10\ \mathrm{MHz}}\approx 2.5\times 10^{-3},\tag{15a}\]

which is consistent with the AD8606 typical closed-loop gain error of about 0.2 \% (\(=2\times 10^{-3}\)). Because the CDC excitation is a 25 kHz square-like drive whose spectrum contains harmonics, the sinusoidal mismatch model (13) understates \(|\varepsilon_{A}|\) at those harmonics, so (15a) is only a fundamental-frequency optimistic estimate. Even so, \(2.5\times 10^{-3}\) already sits about one decade above the Far threshold \(|\varepsilon_{A}|\lesssim 2\times 10^{-4}\) derived from (14) and the 20\(\rightarrow\)10 mm increment. The present follower choice may therefore sit outside the Far mismatch gate. Paths that restore margin include a composite/faster shield buffer, a CDC with lower excitation or tighter native SHLD accuracy, or relying on the CDC's native SHLD driver rather than an external op-amp; these remain design options to be verified on the bench, not claims of this paper.

\begin{figure}
\centering
\includegraphics{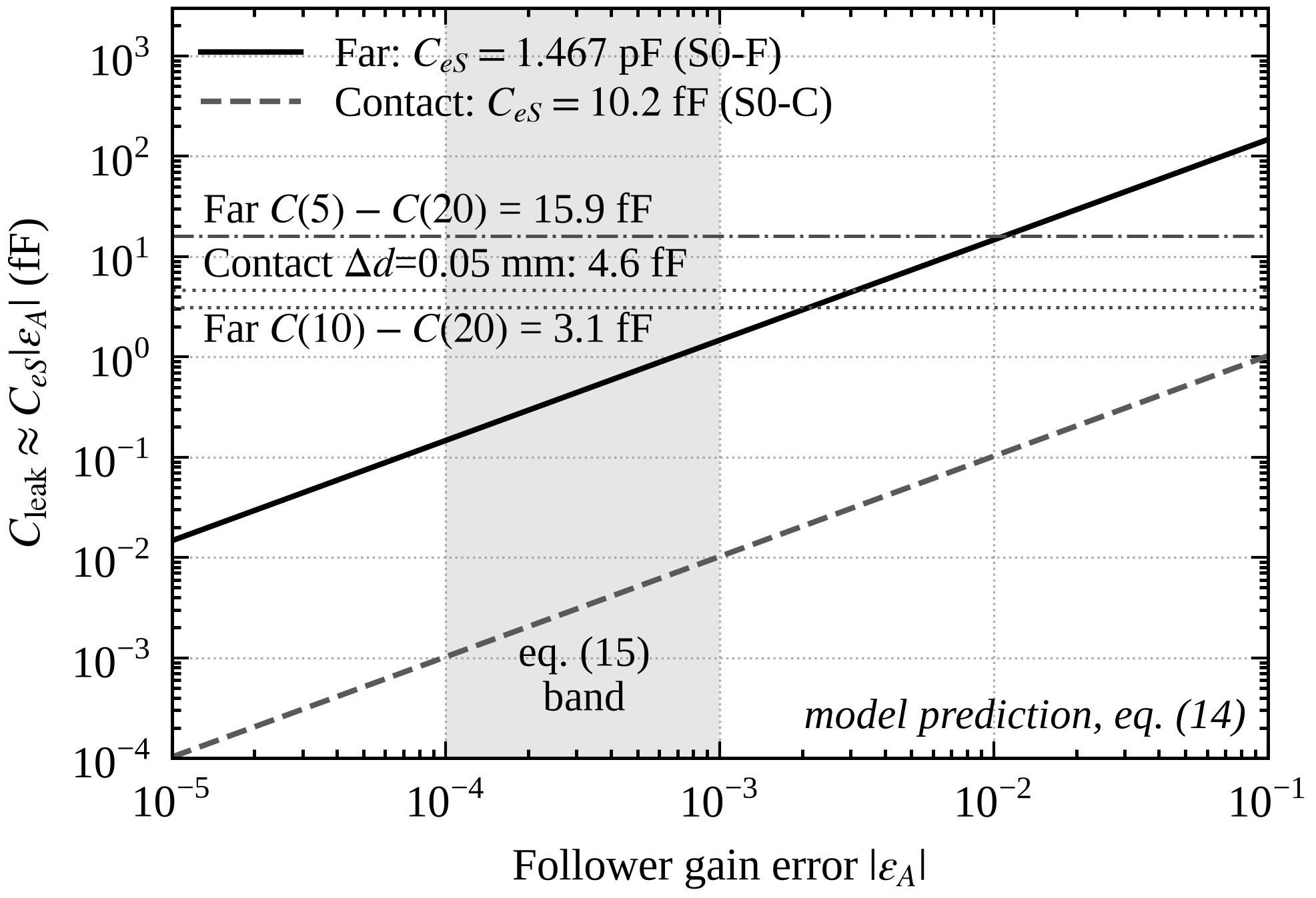}
\caption{Fig. 5}
\end{figure}

\emph{Fig. 5.} Model prediction of the mismatch leakage (14) with FEM-extracted \(C_{eS}\) for the Far (1.467 pF) and Contact (10.2 fF) operators, compared with FEM signal increments. Grey band: the \(|\varepsilon_{A}|\) range of (15).

\subsection{IV. Far-Field Proximity: Image Estimation and Inversion Limits}\label{iv.-far-field-proximity-image-estimation-and-inversion-limits}

\subsubsection{A. Magnitude of the Image-Charge Estimate}\label{a.-magnitude-of-the-image-charge-estimate}

Approximate the target as a grounded conducting sphere of radius \(R\) whose centre is at a distance \(h\) from the electrode plane; the image of a point charge \(q\) in the sphere is \(q'=-qR/r\), with \(r\) the charge-to-centre distance. The dominant term under the small-electrode, distant-sphere approximation is

\[c_{eT}(h)\sim 4\pi\varepsilon_{0}R\cdot\frac{R}{2R+h}\cdot\eta,\tag{16}\]

where \(\eta\) \textless{} 1 accounts for the shield cutting off the lower hemisphere of the field and for the electrode not being a point charge. The role of (16) is to give orders of magnitude and monotonicity, not a distance formula. For the dual-array FEM probe (\(R\) = 5 mm), the simulated increment from 20 mm to 10 mm is only 3.1 fF (Section VIII-B). Backing out \(\eta\) from (16) as

\[\eta \approx \frac{\Delta C_{\mathrm{FEM}}}{(4\pi\varepsilon_{0}R)\,\Delta\!\left(\frac{R}{2R+h}\right)}\approx \frac{3.1\ \mathrm{fF}}{(4\pi\varepsilon_{0}R)\bigl[R/(2R+10\ \mathrm{mm})-R/(2R+20\ \mathrm{mm})\bigr]}\approx 0.07,\tag{16a}\]

shows that (16) over-predicts the bare image-charge increment by more than an order of magnitude unless \(\eta\) is reduced to \(O(10^{-2})\). In this geometry (16) is therefore primarily order-preserving and not quantitative without an FEM-calibrated \(\eta\); shield truncation of field lines toward the phalanx is the leading candidate for the small \(\eta\). Substituting the same \(\eta\approx 0.07\) back into (16) for a finger-scale pad (\(R\) = 8--12 mm, \(h\): 20\(\rightarrow\)10 mm) gives \(\Delta c_{eT}\sim \eta\,(4\pi\varepsilon_{0}R)\,\Delta\!\bigl(R/(2R+h)\bigr)\approx 5\)--\(8\ \mathrm{fF}\) --- single-digit femtofarads, the same order as the 1--10 fF rms noise floor of typical CDCs. That supports treating Far as a decision / threshold channel rather than a precision ranger; the bare ``tens of femtofarads'' estimate without the FEM-calibrated \(\eta\) is not quantitative for this stack.

\subsubsection{B. Ill-Conditioning of Engineering Inversion}\label{b.-ill-conditioning-of-engineering-inversion}

Within a limited window a rational function can absorb the equivalent electrode depth and the skin thickness,

\[C_{\mathrm{obj}}(h)=C_{\infty}+\frac{a}{h+h_{0}}+\frac{b}{(h+h_{0})^{2}},\tag{17}\]

\[\Delta C(h)=C_{\mathrm{meas}}(h)-C_{\mathrm{base}},\qquad h=\mathcal{F}(\Delta C).\tag{18}\]

The baseline \(C_{base}\) must be obtained by slow tracking in a target-free window, otherwise temperature-induced drift of the baseline and micro-expansion will be read as distance. The derivative of the inversion \(h\) = \(F\)(\(\Delta\)\(C\)) with respect to \(\Delta\)\(C\) shrinks rapidly at long range: the same capacitance noise maps to ever larger distance errors. The proximity channel is therefore mechanistically ``interpolable at close range, decision-only at long range''. Stating a stable detection distance as a hard specification requires simultaneously giving the target's electrical properties, the skin and phalanx conditions, and the threshold-to-noise ratio; this paper does not preset these numbers.

\subsubsection{C. Spatial Filtering Implied by Electrode Paralleling}\label{c.-spatial-filtering-implied-by-electrode-paralleling}

Far mode parallels the 16 electrodes, which is equivalent to a spatial low-pass over the array and reduces the channel count. The far-field signal is not amplified \(N\)-fold as a parallel-plate reading would be, because every electrode sees the same highly correlated fringing field rather than \(N\) independent parallel plates. The practical gain is a few-fold, not \(N\)-fold, arising from the increased collection area and the reduced relative noise, while \(C_{para}\) rises as well. Paralleling improves threshold detection; it cannot turn the ill-posed ranging of (16) into well-posed ranging. Paralleling the four corners preserves a weak quadrant bias and is a compromise between scalar detection and centroid estimation. In the dual-array stack the paralleled layer is the outer array L2, one skin thickness closer to the target than an electrode placed under the compressible layer.

\subsection{V. Reconfigurable Observation Operators, Centroid and Crosstalk}\label{v.-reconfigurable-observation-operators-centroid-and-crosstalk}

\subsubsection{A. Three Operators for Two Physical Regimes}\label{a.-three-operators-for-two-physical-regimes}

Let the array pitch be \(p\); the sensitive area is then about 4\(p\) across. What correlates with distance is the footprint width, not a nominal range:

\begin{itemize}
\tightlist
\item
  Far (far field): footprint wider than the array; the 16 L2 electrodes are paralleled onto one CDC input with L4 as active shield; the output is a scalar \(\Delta C_{Far}\) (and its rate of change) used for presence, coarse distance and weak directionality.
\item
  Mid (transition): a neighbouring L2 pair is connected as a hardware differential (\(E_{11}\) \(\rightarrow\) CIN\_A, \(E_{12}\) \(\rightarrow\) CIN\_B), all other L2 and all L4 electrodes on the shield; the output is \(C_{diff}=C_{A}-C_{B}\), whose sign gives the lateral offset. A software difference of two sequential single-ended readings is excluded: it doubles the slot count and subtracts two large, separately drifting baselines instead of cancelling the common mode in the charge domain.
\item
  Contact (near field): all L2 electrodes grounded; the 16 L4 electrodes are scanned point-wise, each with the other 15 on the shield; the output is \(\Delta C_{i}\) and the pressure distribution; the footprint shrinks to one or two pixels.
\end{itemize}

Mode switching should be triggered by the magnitude of the reading relative to its baseline, not by host polling. Entry into Mid uses a dual gate --- a proximity condition and a differential-magnitude condition \(|C_{diff}|\geq\theta_{diff}\) --- while a Far ``pre-arm'' distance only raises the sampling rate or a quality flag and does not hard-switch the routing to Contact. Switching changes the neighbour potentials \(V_{k}\) in (6), so the three operators are not three readings of the same Maxwell matrix but three different sets of boundary conditions. This is the same idea as the dual-mode sensor with reconfigurable electrode interconnection {[}8{]} and the three-mode array that selects its effective aperture with a shield layer {[}15{]}: spatial resolution and detection depth cannot both be optimal under fixed boundary conditions.

\subsubsection{B. Centroid and Its Unbiasedness Conditions}\label{b.-centroid-and-its-unbiasedness-conditions}

For the rectified responses \(\Delta C_{i}^{+}\), the centroid is

\[x_{c}=\frac{\sum_{i}x_{i}\Delta C_{i}^{+}}{\sum_{i}\Delta C_{i}^{+}},\qquad y_{c}=\frac{\sum_{i}y_{i}\Delta C_{i}^{+}}{\sum_{i}\Delta C_{i}^{+}}.\tag{19}\]

Unbiasedness requires a symmetric point-spread function, a footprint not truncated by the array boundary, and no negative sidelobes. In the far field the footprint is wider than the pitch and the centroid is a super-resolution estimate; once the contact patch shrinks, the centroid degenerates to the geometric centre of the occupied pixels and the resolution is set by the pitch. When the target approaches the edge, the centroid shifts inward; this is truncation bias and cannot be removed by any higher-order fit.

\subsubsection{C. Crosstalk Matrix Depends on Switch State}\label{c.-crosstalk-matrix-depends-on-switch-state}

Between the raw observation and the true pixel responses stands

\[\Delta C_{\mathrm{raw}}=M\,\Delta C_{\mathrm{true}}+n.\tag{20}\]

The matrix \(M\) is close to the identity, and its off-diagonal entries come from residual mutual capacitance, MUX leakage and insufficient settling. \(M\) must be calibrated with the same neighbour-electrode potentials as in operation. At run time one can restore the true responses with \(M^{-1}\), or simply subtract the neighbour terms to avoid amplifying noise through inversion. A mutual-capacitance reading (driving rows, sensing columns) is insensitive to the remote ground, carries the opposite sign to self-capacitance, and has a smaller far-field term; it is a candidate for contact-distribution imaging, not the main path for bone-mounted proximity. Under the circ readout of (6b) the mutual terms between the active pad and its shield-tied neighbours drop out identically, so the electrical part of \(M\) reduces to the change of each neighbour's own capacitance to ground; Section VIII-E observes this residual at about 1 fF at the current-mesh convergence level in the contact operator. The mechanical part of \(M\) --- compression of neighbouring L3 cells under a single-pixel load --- is not an electrostatic quantity and is not captured by (6a)--(6c).

\subsection{VI. TDM Settling Time: A Scaling Constraint From Fingertip to Large Area}\label{vi.-tdm-settling-time-a-scaling-constraint-from-fingertip-to-large-area}

\subsubsection{A. First-Order Settling Model}\label{a.-first-order-settling-model}

Time-division multiplexing reduces the number of op-amps to a shared back-end, at the price that after every switching event the channel must settle to a relative error of the same order as (15). The load seen by the follower and the time constant are

\[C_{L}=C_{SG}+C_{\mathrm{MUX}}+C_{\mathrm{trace}}+C_{\mathrm{in}},\qquad \tau=(R_{\mathrm{out}}+R_{\mathrm{on}})C_{L}.\tag{21}\]

and the first-order residual is

\[e(t)=e^{-t/\tau}\,\Delta V_{\mathrm{step}}.\tag{22}\]

Femtofarad-level measurement cannot be scheduled by \(t\) = 3\(\tau\) (10\% residual). For a required settling error \(\varepsilon_{set}\),

\[t_{\mathrm{set}}=-\tau\ln\varepsilon_{\mathrm{set}}=N_{\tau}\,\tau.\tag{23}\]

Thus \(t_{set}\) \(\approx\) 6.9\(\tau\) for \(\varepsilon_{set}\) = 10\(^{-3}\) and \(t_{set}\) \(\approx\) 9.2\(\tau\) for \(\varepsilon_{set}\) = 10\(^{-4}\). The inter-channel crosstalk ratio caused by insufficient settling is on the order of exp(-\(t_{set}\)/\(\tau\)), appearing as a gain that depends on the previous channel's voltage. The frame rate is

\[T_{\mathrm{slot}}=t_{\mathrm{sw}}+t_{\mathrm{set},s}+t_{\mathrm{set},e}+T_{\mathrm{conv}},\qquad f_{\mathrm{frame}}=\frac{1}{N_{\mathrm{ch}}T_{\mathrm{slot}}}.\tag{24}\]

At fingertip scale \(C_{L}\) is about 10--40 pF; with \(R_{out}\) + \(R_{on}\) of tens of ohms, \(\tau\) is on the order of nanoseconds, settling is negligible, and the frame rate is set by \(T_{conv}\) and \(t_{sw}\). At palm scale \(C_{L}\) rises to 50--150 pF, and a single zone of a full arm can reach 200--500 pF; when the same MUX scans hundreds of points, the frame period falls outside a 10 ms safety cycle. ``Zoned followers + Far-mode paralleling + region-of-interest scanning'' is therefore not a product strategy but a corollary of (21)--(24) after area scaling, consistent with the conclusion that large-area tactile skins must solve wiring and multiplexing {[}21{]}.

\subsubsection{B. Implementation in the Dual-Array Readout}\label{b.-implementation-in-the-dual-array-readout}

Fig. 6 shows the readout chain that realises Table I. A single CDC (FDC1004 {[}22{]}, I\textsuperscript{2}C address 0x50) provides two measurement inputs, CIN1 \(\rightarrow\) CIN\_A and CIN2 \(\rightarrow\) CIN\_B, and a shield reference SHLD; CIN3/CIN4 are not used on the TDM path. The shield reference drives a dual follower (Analog Devices AD8606) whose outputs SHLD\_L2 and SHLD\_L4 serve the two arrays separately, so that the follower load of each layer can be examined on its own. Four 8 \(\times\) 8 analog crosspoint switches (ADG2188, I\textsuperscript{2}C 0x70--0x73) connect the 32 electrode nodes to four logical X buses: X0 = GND, X1 = CIN\_A, X2 = CIN\_B, X3 = SHLD (SHLD\_L2 on U4/U5 for L2, SHLD\_L4 on U7/U8 for L4); X4--X7 are reserved and tied to ground. The GND bus is indispensable --- it implements the Faraday screen of (6c) --- and is therefore assigned to X0, which is also the safe idle state after reset; an earlier single-layer routing that used X0 as the guard bus is not compatible.

\begin{figure}
\centering
\includegraphics{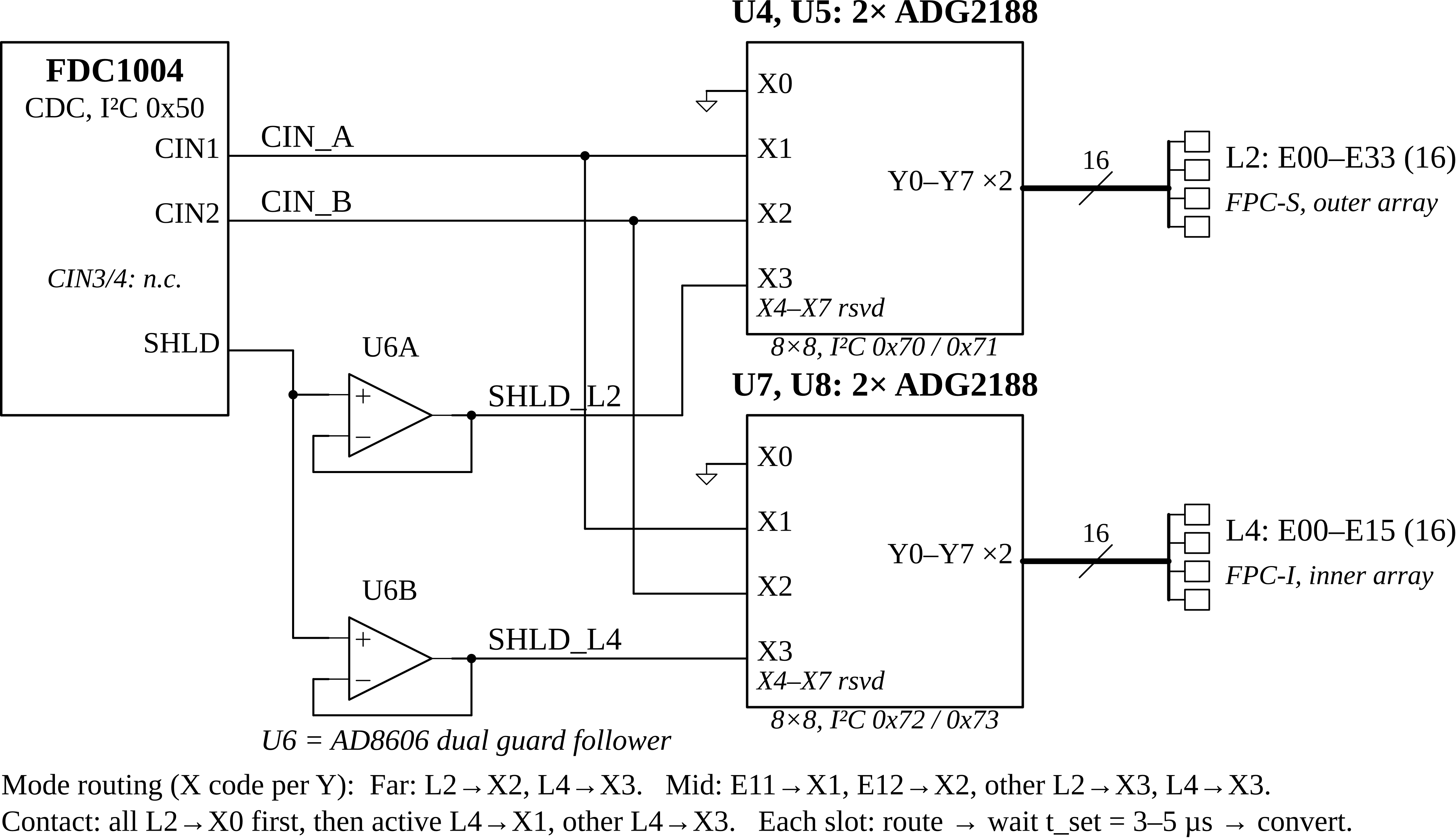}
\caption{Fig. 6}
\end{figure}

\emph{Fig. 6.} Readout chain of the dual-array stack: CDC with two inputs, dual guard follower for SHLD\_L2/SHLD\_L4, and 4 \(\times\) ADG2188 crosspoints mapping the X buses (X0 GND, X1 CIN\_A, X2 CIN\_B, X3 SHLD) onto the 16 L2 and 16 L4 electrodes. The bottom lines list the per-mode routing and the slot sequence. Drawn in KiCad; decoupling, ESD and series elements are omitted.

Mode changes are executed as atomic sequences (Fig. 7). Entering Contact, all L2 electrodes are first connected to GND, and only then is one L4 electrode connected to CIN\_A with the other 15 on SHLD; leaving Contact, the order is reversed. This prevents a window in which an L4 electrode measures through an incomplete screen, and a window in which L2 and L4 are simultaneously on sensitive nodes. The routing decoder rejects illegal combinations --- any L2 on CIN while another L2 is on GND; Contact with any L2 not on GND; Far or Mid with any L4 on CIN; a conversion started before the settling interval --- and keeps the last legal table.

\begin{figure}
\centering
\includegraphics{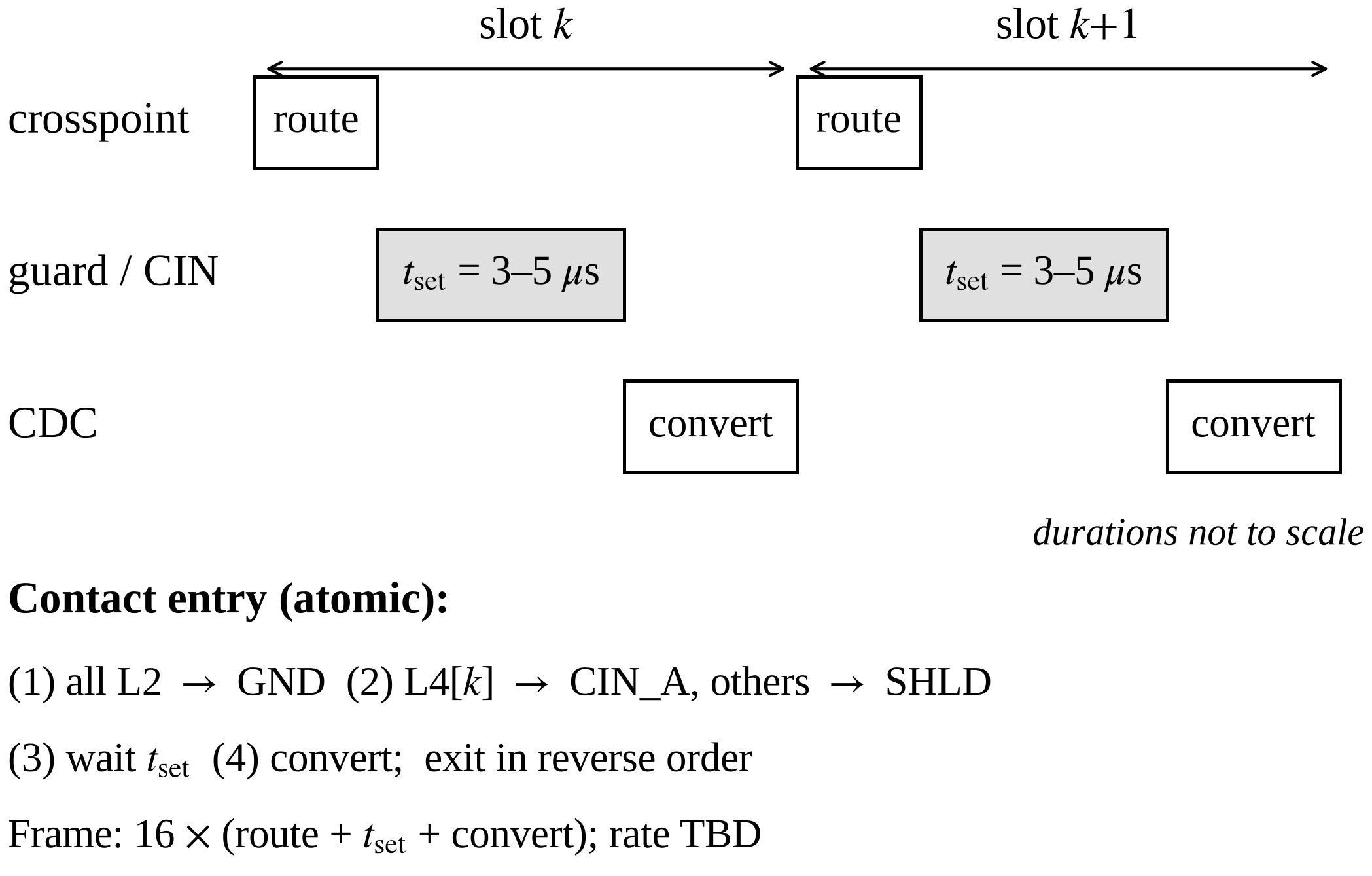}
\caption{Fig. 7}
\end{figure}

\emph{Fig. 7.} Slot sequence of the TDM readout and the atomic Contact entry order. Each slot comprises crosspoint routing, a hard wait \(t_{set}\) = 3--5 \(\mu\)s and one CDC conversion; a Contact frame comprises 16 slots. Durations are not to scale; the frame rate is to be determined on the bench (matplotlib schematic).

The settling interval is locked at \(t_{set}\) = 3--5 \(\mu\)s after every change of any node role. This value is an engineering lock, not a result of (23). The FEM baselines give the sensor-side part of \(C_{L}\) --- the self-capacitance of the shield node, 17.1 pF in the Far configuration and 16.0 pF in the Contact configuration --- in line with the 10--40 pF fingertip estimate above. With a nominal \(R_{out}+R_{on}\) = 50 \(\Omega\), (21)--(23) give \(\tau\) \(\approx\) 0.85 ns and \(t_{set}\) \(\approx\) 8 ns for \(\varepsilon_{set}=10^{-4}\), and the first-order model stays below 1 \(\mu\)s up to \(C_{L}\) \(\approx\) 2 nF (Fig. 8). The 3--5 \(\mu\)s window therefore budgets effects outside (21)--(23): crosspoint charge injection, follower stability and slewing with a capacitive load (in Mid mode up to about 30 crosspoint outputs hang on the shield buses), interconnect parasitics between the flexible circuits and the driver board, and CDC input recovery. A sub-microsecond value derived from the RC term alone was considered and withdrawn; any reduction of \(t_{set}\) requires a bench measurement of the injection transient. With 16 Contact slots, the 3--5 \(\mu\)s window contributes 48--80 \(\mu\)s per frame --- negligible beside the CDC conversion time. The FDC1004 supports 100/200/400 samples/s {[}22{]}. At 400 S/s, \(T_{conv}\approx 1/400=2.5\ \mathrm{ms}\), so a 16-slot Contact frame lasts about \(16\times 2.5=40\ \mathrm{ms}\) (\(\approx 25\ \mathrm{Hz}\)); at the 100 S/s setting used for finest capacitance resolution, the same 16-slot frame lasts about \(16\times 10=160\ \mathrm{ms}\) (\(\approx 6\ \mathrm{Hz}\)). Both already exceed the 10 ms safety-cycle order mentioned after (24). Frame rate and resolution are therefore a trade-off: for femtofarad-level Contact measurements the practically usable rate on this CDC is closer to \textasciitilde6 Hz than to the 25 Hz ceiling. Thus \(T_{conv}\) in (24), not settling, is the frame-rate bottleneck for the present CDC choice; the frame rate itself is not a measured claim. Faster scanning would require using the unused CIN3/CIN4 channels for limited parallelism, reducing Contact resolution (fewer pads per frame), or a faster CDC. The 50 \(\Omega\) value is a nominal placeholder; the switch on-resistance and capacitances are not taken from a datasheet in this paper.

\begin{figure}
\centering
\includegraphics{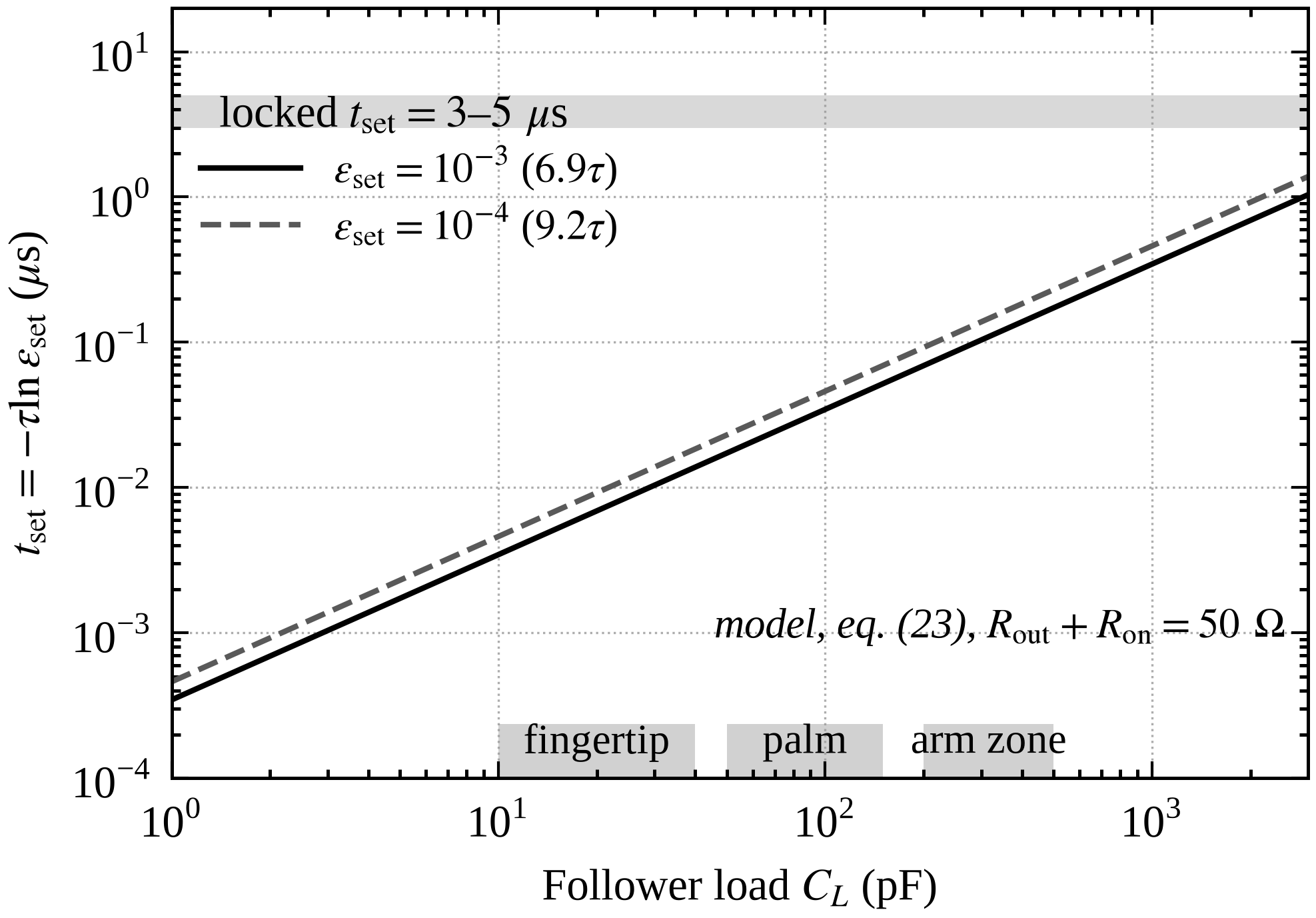}
\caption{Fig. 8}
\end{figure}

\emph{Fig. 8.} Model prediction of the first-order settling time (23) versus follower load for \(R_{out}+R_{on}\) = 50 \(\Omega\) (nominal), with the locked 3--5 \(\mu\)s window and the \(C_{L}\) ranges of Section VI-A. The gap between the curves and the window is the margin reserved for non-RC effects.

\subsection{VII. Contact Regime: Finite Compression, Material Cross-Sensitivity and the Unobservable Tangential Component}\label{vii.-contact-regime-finite-compression-material-cross-sensitivity-and-the-unobservable-tangential-component}

\subsubsection{A. Breakdown of the Linear Parallel-Plate Form}\label{a.-breakdown-of-the-linear-parallel-plate-form}

During contact \(d\) = \(d_{0}\) - \(\Delta\)\(d\), and from (1),

\[\frac{\Delta C}{C_{0}}=\frac{\varepsilon_{r}}{\varepsilon_{r0}}\,\frac{d_{0}}{d_{0}-\Delta d}-1.\tag{25}\]

For small deformation with \(\varepsilon_{r}\) unchanged, \(\Delta\)\(C\)/\(C_{0}\) \(\approx\) \(\Delta\)\(d\)/\(d_{0}\). When the local compression reaches tenths of the skin thickness, neither the higher-order terms nor a possible change of \(\varepsilon_{r}\) can be ignored, and a linear calibration cannot be extrapolated to full scale. Equation (25) also assumes that the whole of \(C_{0}\) is the compressible plate. In a real stack only part of \(C_{0}\) is: with \(C_{0}=C_{\mathrm{fix}}+C_{\mathrm{gap}}\), the relative change is diluted by \(C_{\mathrm{gap}}/C_{0}\). For the dual-array stack the FEM fit of Section VIII-D attributes about 74 fF of the 1.08 pF base to the \(1/(d_{0}-\Delta d)\) term; the fixed part is dominated by the coupling of the active L4 pad to the grounded L5 pour across the thin L4--L5 dielectric, for which (1) gives about 0.8 pF (0.075 mm, \(\varepsilon_{r}\) = 3.5, 1.96 mm\textsuperscript{2}). Tying L5 to the shield instead of ground in Contact mode would remove this term from (6a) and raise the relative sensitivity, at the price of raising \(C_{eS}\) in Contact mode to the same order and thereby re-introducing the mismatch constraint of (14). This trade-off is an analytical prediction; it has not been simulated.

\subsubsection{B. Force--Capacitance Mapping}\label{b.-forcecapacitance-mapping}

The local contact of a silicone skin is close to Hertzian,

\[F_{n}=k_{\mathrm{eff}}(F_{n})\,\Delta d,\qquad k_{\mathrm{eff}}\propto E^{*}\sqrt{R\,\Delta d}.\tag{26}\]

The effective stiffness grows with compression, and the capacitance--force relation is ``geometric nonlinearity \(\times\) material nonlinearity''. It can be written as

\[F_{n}=\mathcal{G}\!\left(\textstyle\sum_{i}\Delta C_{i}\right),\tag{27}\]

where the mapping \(G(\cdot)\) (denoted by script G in (27)) is determined by loading experiments; a polynomial is only a parameterisation, not a constitutive law. The unloaded baseline \(C_{0}\) must be subtracted at the input of \(G\), otherwise slow drift is read as force. The pressure distribution can be output pixel-wise as \(\Delta\)\(C_{i}\); it is a capacitance image and only approximates a pressure image after calibration. In the dual-array stack (27) factors into an electrostatic part \(C(\Delta d)\), which FEM provides (Section VIII-D), and a mechanical part \(\Delta d(F_{n})\), which requires hyperelastic simulation or loading data and is not available; \(\mathcal{G}\) is therefore not determined in this paper. The coarse \(C(F)\) conversion reported for the earlier single-layer stack is withdrawn: it was derived for the open-field path discussed next.

\subsubsection{C. Tangential: Centroid Azimuth Is Not Tangential Force}\label{c.-tangential-centroid-azimuth-is-not-tangential-force}

An azimuth can be defined from the centroid,

\[\theta=\mathrm{atan2}(y_{c},x_{c}).\tag{28}\]

Quantising \(\theta\) into a finite set of directions only states the offset direction of the contact patch relative to the array centre. The same shear force produces different \(\theta\) under different normal preloads, friction coefficients and contact-patch areas; the present structure has no independent tangential capacitance channel, so \(\theta\) is not an observer of tangential force. Continuous tangential force requires staggered electrode pairs, four-quadrant differential readout or a sliding-dielectric structure {[}13{]}, {[}16{]}, which belong to a different class of observation models. In the dual-array stack, the centroid of (19) and the azimuth of (28) are computed on the L4 contact image rather than on the outward-looking L2 layer.

\subsubsection{D. Material Cross-Sensitivity of a Single-Layer Stack and Its Removal by a Faraday Screen}\label{d.-material-cross-sensitivity-of-a-single-layer-stack-and-its-removal-by-a-faraday-screen}

In a single-layer stack the electrodes that look outward for proximity also measure force by the compression of the skin above them. During contact the target is then a conductor (or dielectric) in the immediate field of the active electrode, and the reading change is

\[\Delta C_{\mathrm{meas}}=\Delta C_{\mathrm{force}}(\Delta d)+\Delta C_{\mathrm{dielectric}}(\varepsilon_{\mathrm{target}},\,\sigma_{\mathrm{target}},\,A_{\mathrm{contact}}),\tag{25a}\]

where the second term depends on the permittivity, conductivity and grounding of the touched object and on the contact area. A grounded metal probe and an insulating probe pressing with the same force therefore produce different readings, and no single map \(\mathcal{G}\) in (27) can be material-independent. Estimating \(\varepsilon_{\mathrm{target}}\) before contact and selecting among several maps is possible in principle but depends on humidity and on the grounding state of the object; it is not a substitute for removing the second term physically. The single-layer contact FEM of Section VIII-A --- a grounded metal probe on a uniformly thinned skin --- describes exactly this open-field path; it is kept as a counter-example and is not used as contact calibration data.

The dual-array stack removes the second term by construction. In Contact mode the outer array L2 is grounded (Table I), and the active L4 pad reads (6c): the target field ends on L2, and \(\varepsilon_{\mathrm{target}}\) can enter only through the residual \(c_{iT}^{\mathrm{res}}\). The figures of merit are the material leakage and its ratio to the force signal,

\[\Delta C_{\mathrm{leak}}(\Delta d)=C_{\mathrm{metal}}(\Delta d)-C_{\mathrm{nylon}}(\Delta d),\qquad \varepsilon_{\mathrm{xtalk}}=\frac{\Delta C_{\mathrm{leak}}(\Delta d)}{C_{\mathrm{metal}}(\Delta d)-C_{\mathrm{metal}}(0)}.\tag{25b}\]

The electrostatic part of (25b) is evaluated by FEM in Section VIII-D. The system-level acceptance --- the same 5 N load at the same position with an insulating nylon probe and a grounded metal probe yielding \(|\Delta F_{n}|\leq0.1\) N --- additionally needs \(\mathcal{G}\) and remains open.

\subsection{VIII. Numerical Validation (FEM Simulation)}\label{viii.-numerical-validation-fem-simulation}

Two electrostatic capacitance-matrix campaigns are reported. Both use ANSYS Maxwell 3D Electrostatic driven by PyAEDT (the single-layer campaign used release 2024 R1). All simulated values in this section are design evidence, not measured acceptance criteria.

\emph{Single-layer campaign (V1, historical).} The observable was the driven-shield reading \(C_{circ}\) = \(C\)(\(V_{e}\), \(V_{e}\)) + \(C\)(\(V_{e}\), \(V_{Shield}\)). The fixtures were: for the Far case, a 20 mm-diameter conducting sphere with the sensor skin-covered and bone-mounted, sweeping \(h\), with \(\Delta\)\(C\) = \(C\)(10 mm) - \(C\)(20 mm); for the Mid case, the \(h\) \(\times\) \(e_{x}\) matrix (25 points) of the neighbouring pads E11--E12; for the Contact case, the single electrode E11 with a uniform thinning \(\delta\) of the outer skin (numerical residual gap 0.05 mm). The stack-up was Pad/Skin/PDMS/Shield/Bone with an outer-skin Dk of 3.

\emph{Dual-array campaign (V2, frozen 25 Sep 2026).} The geometry is that of Figs. 2 and 3 (L1 0.8 mm, \(\varepsilon_{r}\) 3.0; L3 0.6 mm, \(\varepsilon_{r}\) 2.7; L4--L5 dielectric 0.075 mm, \(\varepsilon_{r}\) 3.5; phalanx 0.5 mm; residual air gap 0.05 mm; pad 1.4 mm, pitch 2.0 mm). The probes are a grounded metal sphere (\(R\) = 5 mm), a floating plastic sphere (\(\varepsilon_{r}\) = 3.0) and a floating nylon sphere (\(\varepsilon_{r}\) = 3.5). Node potentials follow Table I, with 1 V on CIN and SHLD and 0 V on GND, L5, phalanx and metal probe. L2 is modelled as one ideal equipotential sheet (a continuous grounded conductor), not as sixteen discrete pads at 0 V: inter-pad gaps, seams and the finite impedance of the L2 ground return are omitted. A real 16-pad Faraday screen would leave fringing paths for \(c_{iT}^{\mathrm{res}}\), so the Contact material-leakage figures below are optimistic relative to hardware. The mesh is the coarse ``Track-A'' setting (electrode length 0.35 mm, dielectric 0.5 mm, at most 6 adaptive passes, 0.3 \% energy error), coarser than the 0.25--0.30 mm and 10 passes recommended in the simulation plan. Every Contact solution (7/7 in C1 and C2, 3/3 in C4) and both empty baselines logged an adaptive non-convergence warning while completing normally and exporting the matrix; all Contact and empty-baseline claims below are therefore stated at the convergence level of the current mesh, not as mesh-converged bounds. The C1/C2 matrices print \(C\) to 0.1 fF; the 16-port C4 matrices carry 13 significant digits. Raw diagonal entries (\(C_{kk}\)) are reported as exported; circ values follow (6b) and, where marked ``derived'', were computed in this paper from the exported entries.

\subsubsection{A. Single-Layer Stack: Historical Results Retained as Counter-Example}\label{a.-single-layer-stack-historical-results-retained-as-counter-example}

\emph{Far.} The baseline (16 pads fully paralleled, pad 1.4, Dk 3) gave \(C\)(20 mm) = 153.7 fF and \(\Delta\)\(C\) = +9.4 fF (6.12\%); the high-copper variant (pad 1.9) gave +11.6 fF; high-permittivity outer skins with Dk 6/12/20 gave +11.8/+15.4/+18.8 fF; and the multiplicative combination pad 1.9 \(\times\) Dk20 gave 488.6 fF and +21.3 fF (4.36\%). The unshielded control (16 pads paralleled, no driven shield) read 2206 fF with \(\Delta\)\(C\) of only +4.7 fF (0.21\%), confirming (8) and the statement that the shield cannot be removed. The hollowed, checkerboard, loop-ring and focusing-wall topologies all locked \(\Delta\)\(C\)/\(C\) at about 6\%. Because the absolute upper bound of +21.3 fF fell below the original absolute threshold (about 40 fF), the Far criterion was revised to a rate-of-change threshold \textbar d\(C\)/d\(t\)\textbar{} \(\geq\) \(\theta_{rate}\) (equivalent to about 3--5 fF), with the absolute \(\Delta\)\(C\) kept only as a quality bit; \(\theta_{rate}\) is to be set as a multiple (5--8) of the measured standard deviation of d\(C\) and has not been measured. The single-layer Far lock (pad \(\approx\) 1.9 mm, engineering Dk 6--12) is not part of the dual-array FEM baseline (pad 1.4 mm, \(\varepsilon_{r}\) 3.0); whether it is adopted is decided by the Far non-regression test (Section VIII-F). Table II compares the simulated schemes.

\textbf{TABLE II} Far-Mode FEM Scheme Comparison (Single-Layer Stack, V1)

\begin{longtable}[]{@{}
  >{\raggedright\arraybackslash}p{(\columnwidth - 8\tabcolsep) * \real{0.2000}}
  >{\raggedright\arraybackslash}p{(\columnwidth - 8\tabcolsep) * \real{0.2000}}
  >{\raggedright\arraybackslash}p{(\columnwidth - 8\tabcolsep) * \real{0.2000}}
  >{\raggedright\arraybackslash}p{(\columnwidth - 8\tabcolsep) * \real{0.2000}}
  >{\raggedright\arraybackslash}p{(\columnwidth - 8\tabcolsep) * \real{0.2000}}@{}}
\toprule\noalign{}
\begin{minipage}[b]{\linewidth}\raggedright
Scheme
\end{minipage} & \begin{minipage}[b]{\linewidth}\raggedright
\(C\)(20 mm) (fF)
\end{minipage} & \begin{minipage}[b]{\linewidth}\raggedright
\(\Delta\)\(C\) (fF)
\end{minipage} & \begin{minipage}[b]{\linewidth}\raggedright
\(\Delta\)\(C\)/\(C\)
\end{minipage} & \begin{minipage}[b]{\linewidth}\raggedright
Remark
\end{minipage} \\
\midrule\noalign{}
\endhead
\bottomrule\noalign{}
\endlastfoot
All-parallel, pad 1.4, Dk 3 & 153.7 & +9.4 & 6.12\% & Baseline \\
High-copper, pad 1.9 & 197.9 & +11.6 & 5.86\% & --- \\
Outer skin Dk 6/12/20 & 216/319/421 & +11.8/+15.4/+18.8 & 5.45/4.83/4.47\% & --- \\
Pad 1.9 \(\times\) Dk20 & 488.6 & +21.3 & 4.36\% & Absolute upper bound \\
All-parallel, unshielded & 2206 & +4.7 & 0.21\% & Shield cannot be removed \\
Centre 4 + outer 12 shield wall & \textasciitilde22 & \textasciitilde+2 & \textasciitilde10\% & --- \\
Mutual-capacitance adjacent peak & --- & \textless0.5 & --- & --- \\
Adjacent-pad differential mode & --- & \(\leq\)0.03 & --- & --- \\
\end{longtable}

\emph{Mid.} In the M1 batch (\(h\) \(\in\) \{10, 8, 5, 3, 1\} mm \(\times\) \(e_{x}\) \(\in\) \{0, \(\pm\)2, \(\pm\)4\} mm) the differential reading \(C_{diff}\) = \(C_{a}\) - \(C_{b}\) was approximately odd-symmetric (Table III). At \(h\) = 10/8 mm it was \(\approx\) 0; at \(h\) \(\approx\) 5 mm the eccentric differential first exceeded 0.1--0.2 fF; at \(h\) = 3--1 mm it grew from the sub-fF level to several fF (\(\pm\)4.30 fF at \(e_{x}\) = \(\pm\)4, \(h\) = 1). On this basis the single-layer design used the distance bands Far 20--5 mm, Mid 5--1 mm and Contact within about 1 mm, with a dual gate prox \(\leq\) 5 mm and \textbar{}\(C_{diff}\)\textbar{} \(\geq\) 0.1 fF. These bands are not carried over to the dual-array stack (Section VIII-C).

\textbf{TABLE III} Mid-Mode Differential Capacitance \(C_{diff}\) (fF), Single-Layer Stack (V1)

\begin{longtable}[]{@{}llllll@{}}
\toprule\noalign{}
\(e_{x}\) ~\(h\) (mm) & 10 & 8 & 5 & 3 & 1 \\
\midrule\noalign{}
\endhead
\bottomrule\noalign{}
\endlastfoot
0 & 0.02 & 0.00 & 0.01 & 0.02 & 0.10 \\
\(\pm\)2 & \(\pm\)0.03 & \(\pm\)0.04 & \(\pm\)0.15 & \(\pm\)0.55 & \(\pm\)3.85 \\
\(\pm\)4 & \(\pm\)0.02 & \(\pm\)0.07 & \(\pm\)0.24 & \(\pm\)0.79 & \(\pm\)4.30 \\
\end{longtable}

\emph{Contact (open field).} In the C1 batch, \(\delta\) = 0 \(\rightarrow\) 71.09 fF; \(\delta\) = 0.05 mm \(\rightarrow\) 74.82 fF (\(\Delta\)\(C\) \(\approx\) +3.73 fF); \(\delta\) = 0.2 mm \(\rightarrow\) 83.04 fF, a trend consistent with (25). These values were obtained with a grounded metal probe acting directly on the sensing electrode through the skin, i.e.~on the open-field path of (25a); the reading contains the probe's own coupling and cannot be transferred to other materials. The coarse force conversion previously derived from them is withdrawn (Section VII-B), and the uncompleted points of that batch are not used as calibration input for the dual-array stack.

\subsubsection{B. Dual-Array Far (F1--F3)}\label{b.-dual-array-far-f1f3}

All 9 + 15 + 2 Far points were solved (Table IV, Fig. 9). In the empty domain the paralleled L2 array reads \(C_{\mathrm{CIN_B}}\) = 2.6168 pF; with the metal probe the Far reference is \(C\)(20 mm) = 2.6255 pF, taken from the same domain rather than from the empty baseline because the two differ in domain and mesh. The practical gate is the raw increment \(C\)(5 mm) - \(C\)(20 mm) = 15.9 fF (0.61 \% of the base); the increment from 20 mm to 10 mm is only 3.1 fF (0.12 \%). Evaluated with the circ readout (6b), the base drops to 1157.7 fF and the increments become 17.2 fF and 4.2 fF (1.49 \% and 0.36 \%, derived). The circ base is still large because the L4 shield consists of pads: through the gaps between them the paralleled L2 array keeps a direct coupling to the grounded L5 pour and phalanx (Section III-A).

Lateral offset (F2) at \(h\) = 5 mm lowers the reading monotonically with \(|e_{x}|\) (2.6414 pF at \(e_{x}\) = 0 to 2.6377/2.6381 pF at \(e_{x}\) = \(\mp\)4 mm). At \(h\) = 20 and 10 mm the offset-induced changes are about 1 fF, at the level of the cross-design numerical scatter, so the paralleled Far channel carries no usable lateral information at those distances; a multi-pad Far centroid has not been evaluated. The plastic probe (F3) reads 6.0 fF and 10.5 fF below the metal probe at \(h\) = 20 and 10 mm. Its absolute values also lie below the metal reference at 20 mm, so a cross-design scatter of a few femtofarads cannot be excluded; the plastic--metal difference nevertheless shows an open-field material sensitivity of the same order as the Far signal itself. The dual-array Far numbers cannot be compared directly with Table II: the probe diameter (10 mm versus 20 mm), the stack and the extracted quantity differ. Whether the dual-array Far channel regresses relative to the single-layer lock therefore remains open.

\textbf{TABLE IV} Dual-Array Far Mode, Metal Sphere \(R\) = 5 mm (F1); Circ Values Derived via (6b)

\begin{longtable}[]{@{}
  >{\raggedright\arraybackslash}p{(\columnwidth - 8\tabcolsep) * \real{0.2000}}
  >{\raggedright\arraybackslash}p{(\columnwidth - 8\tabcolsep) * \real{0.2000}}
  >{\raggedright\arraybackslash}p{(\columnwidth - 8\tabcolsep) * \real{0.2000}}
  >{\raggedright\arraybackslash}p{(\columnwidth - 8\tabcolsep) * \real{0.2000}}
  >{\raggedright\arraybackslash}p{(\columnwidth - 8\tabcolsep) * \real{0.2000}}@{}}
\toprule\noalign{}
\begin{minipage}[b]{\linewidth}\raggedright
\(h\) (mm)
\end{minipage} & \begin{minipage}[b]{\linewidth}\raggedright
\(C_{kk}\) raw (pF)
\end{minipage} & \begin{minipage}[b]{\linewidth}\raggedright
\(\Delta\)\(C\) raw (fF)
\end{minipage} & \begin{minipage}[b]{\linewidth}\raggedright
\(C_{\mathrm{circ}}\) (fF)
\end{minipage} & \begin{minipage}[b]{\linewidth}\raggedright
\(\Delta\)\(C_{\mathrm{circ}}\) (fF)
\end{minipage} \\
\midrule\noalign{}
\endhead
\bottomrule\noalign{}
\endlastfoot
0.1 & 2.9285 & 303.0 & 1477.1 & 319.4 \\
1 & 2.7434 & 117.9 & 1282.3 & 124.6 \\
3 & 2.6608 & 35.3 & 1195.8 & 38.1 \\
5 & 2.6414 & 15.9 & 1174.9 & 17.2 \\
8 & 2.6317 & 6.2 & 1164.4 & 6.7 \\
10 & 2.6286 & 3.1 & 1161.9 & 4.2 \\
12 & 2.6280 & 2.5 & 1160.4 & 2.7 \\
15 & 2.6272 & 1.7 & 1159.2 & 1.5 \\
20 & 2.6255 & 0.0 (ref.) & 1157.7 & 0.0 (ref.) \\
F3 plastic, 20 & 2.6195 & -6.0 vs metal & 1151.0 & -6.7 vs metal \\
F3 plastic, 10 & 2.6181 & -10.5 vs metal & 1150.7 & -11.2 vs metal \\
\end{longtable}

\begin{figure}
\centering
\includegraphics{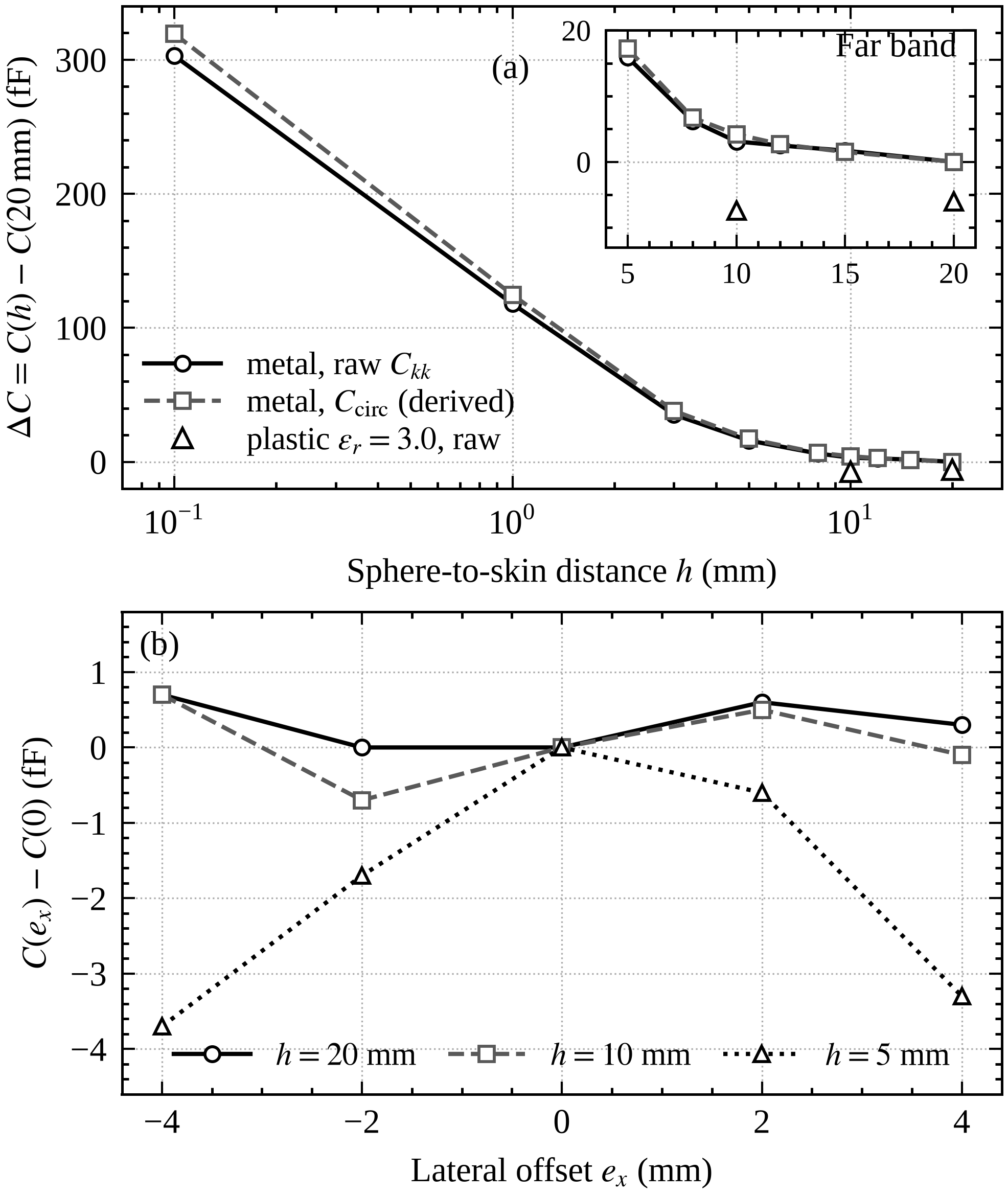}
\caption{Fig. 9}
\end{figure}

\emph{Fig. 9.} Dual-array Far mode (FEM). (a) Increment relative to \(h\) = 20 mm for the metal sphere, raw diagonal and derived circ readout, with the plastic probe (F3); inset: Far band. (b) Offset response \(C(e_{x})-C(0)\) (raw) at \(h\) = 20, 10 and 5 mm (F2).

\subsubsection{C. Dual-Array Mid (M1)}\label{c.-dual-array-mid-m1}

All 26 Mid points were solved. The common-mode condition at zero offset holds (\(|C_{diff}|/C_{cm}\) \(\leq\) 0.46 \% raw, \(\leq\) 0.493 \% circ). Evaluated with raw diagonal entries, the differential signal is only 0.1--1.2 fF and non-monotonic in \(|e_{x}|\) at every \(h\) --- inside the numerical noise band. The cause is the extraction, not the physics: the raw diagonal omits the couplings to the two shield buses, which (6a) shows to be part of what the CDC reads. Recomputing the same matrices with the circ form \(C_{k}=C_{kk}+C_{k,\mathrm{SHLD\_L2}}+C_{k,\mathrm{SHLD\_L4}}\), at no additional solver cost, gives Table V and Fig. 10. At \(h\) = 1 mm the differential reaches 3.1--4.2 fF with the correct sign; at \(h\) = 3 mm it is 0.41--0.60 fF; at \(h\) \(\geq\) 5 mm it stays within \(\pm\)0.25 fF. Strict monotonicity in \(|e_{x}|\) holds at \(h\) = 3 and 8 mm and fails at \(h\) = 1, 5 and 10 mm (at \(h\) = 1 mm, \(|C_{diff}|\) at \(|e_{x}|\) = 4 mm is smaller than at 2 mm). With a 0.3 fF magnitude gate, 8 of 20 non-zero-offset points yield a direction decision (5 of 20 with 0.5 fF); all of them lie at \(h\) \(\leq\) 3 mm and none has the wrong sign.

The Mid information is therefore fused with Far rather than used as an independent band. With \(s_{M}\) the mean \(|C_{diff}|\) at \(|e_{x}|\) = 2 and 4 mm, the gate weight \(w_{M}=\mathrm{clip}\big((s_{M}-0.3\ \mathrm{fF})/(1.0\ \mathrm{fF}-0.3\ \mathrm{fF}),0,1\big)\) and \(w_{F}=1-w_{M}\) give \(w_{M}\) = 0 for \(h\) \(\geq\) 5 mm, 0.23 at 3 mm and 1.0 at 1 mm. Compared with Table III, the dual-array Mid signal at \(h\) = 5 mm is much weaker (circ \(|C_{diff}|\) = 0.003/0.003/0.116 fF at \(|e_{x}|\) = 0/2/4 mm versus 0.01/0.15/0.24 fF); the probe size, the stack and the mesh changed together, and their contributions have not been separated. The single-layer Mid band (5--1 mm) and its 0.1 fF gate are therefore not transferred: in the dual-array stack the Mid direction is usable only for \(h\) \(\leq\) 3 mm, the proximity threshold of the dual gate must be re-calibrated on L2, and a circ extraction is mandatory in firmware.

\textbf{TABLE V} Dual-Array Mid Mode, Circ Differential \(C_{diff}\) (fF), Metal Sphere \(R\) = 5 mm (M1)

\begin{longtable}[]{@{}llllll@{}}
\toprule\noalign{}
\(e_{x}\) (mm) ~\(h\) (mm) & 10 & 8 & 5 & 3 & 1 \\
\midrule\noalign{}
\endhead
\bottomrule\noalign{}
\endlastfoot
-4 & -0.06 & 0.07 & 0.07 & 0.60 & 3.14 \\
-2 & -0.07 & -0.07 & 0.00 & 0.42 & 3.96 \\
0 & 0.17 & 0.01 & 0.00 & 0.36 & 0.15 \\
+2 & 0.03 & 0.10 & 0.00 & -0.41 & -4.20 \\
+4 & -0.18 & -0.25 & -0.16 & -0.42 & -3.10 \\
\(w_{M}\) (gate) & 0 & 0 & 0 & 0.23 & 1.0 \\
\end{longtable}

\begin{figure}
\centering
\includegraphics{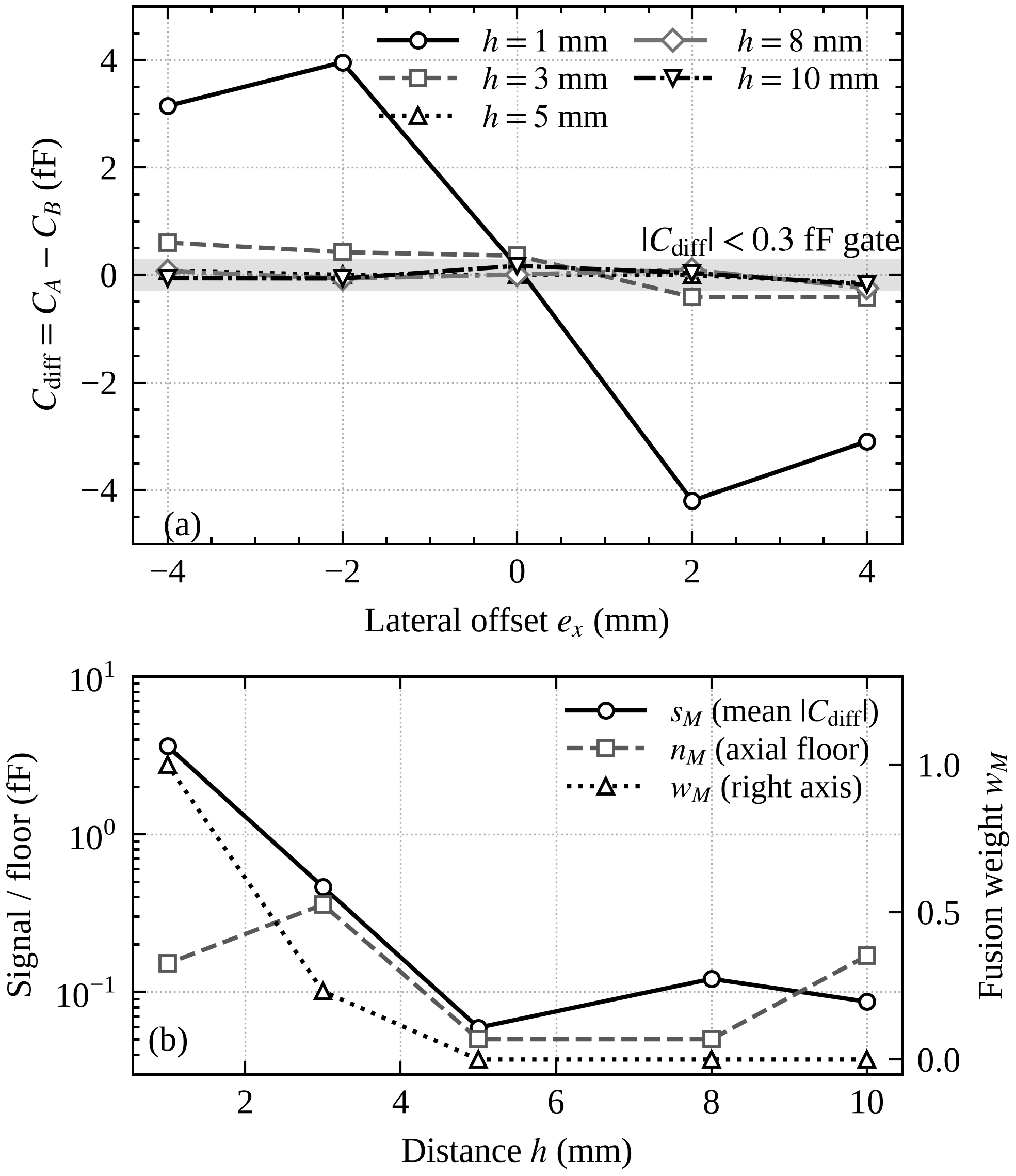}
\caption{Fig. 10}
\end{figure}

\emph{Fig. 10.} Dual-array Mid mode (FEM, circ readout). (a) \(C_{diff}\) versus lateral offset; grey band: 0.3 fF magnitude gate. (b) Differential signal \(s_{M}\), axial floor \(n_{M}\) (\(|C_{diff}|\) at \(e_{x}\) = 0) and fusion weight \(w_{M}\) versus distance.

\subsubsection{D. Dual-Array Contact: Compression Response and Material Independence (C1, C2)}\label{d.-dual-array-contact-compression-response-and-material-independence-c1-c2}

In C1 the grounded metal probe rests 0.05 mm above the skin over \(E_{11}\), all L2 electrodes are grounded, L4{[}\(E_{11}\){]} is on CIN\_A and the other 15 L4 pads are on SHLD; the whole L3 layer is compressed to \(d_{0}-\Delta d\). The response is monotonic at all 7 points (Table VI, Fig. 11): from 1.0769 pF at \(\Delta\)\(d\) = 0 (0.6 fF above the empty-domain value of 1.0763 pF) to 1.4469 pF at \(\Delta\)\(d\) = 0.5 mm, i.e.~4.6 fF at 0.05 mm, 29.8 fF at 0.2 mm and 370.0 fF at 0.5 mm. The sensitivity is strongly nonlinear: the forward interval slope between successive points rises from 92 to 2320 fF/mm, about 25-fold. A single-term fit \(C=a+b/(d_{0}-\Delta d)\) gives \(a\) = 0.99738 pF, \(b\) = 44.618 fF\(\cdot\)mm and \(R^{2}\) = 0.9990, but its residuals (maximum 5.57 fF) have the systematic sign pattern +, +, +, -, -, -, +, showing that the parallel-plate form misses the fringing and series-dielectric contributions. The segmented form developed for the single-layer stack (linear, quadratic and a power law with exponent capped at 3) fits the first five points but leaves residuals of 15.4 fF at 0.4 mm and 151.3 fF at 0.5 mm (rms 57.5 fF, derived). The compression response is therefore implemented as a monotonic look-up table with piecewise-linear or monotone-cubic interpolation rather than as a closed form; this also follows from Section VII-A.

C2 repeats C1 with the floating nylon probe. The two curves coincide within \(|\Delta C_{\mathrm{leak}}|\) \(\leq\) 0.3 fF over the whole range at the print resolution of the exported matrix --- one to three print LSBs --- and the nylon curve is itself monotonic. That 0.3 fF figure is an observation at the present mesh/print resolution, not a proven physical upper bound: Contact C1/C2/C4 and the empty baselines all completed under adaptive non-convergence warnings, and the C4 mirror-pair asymmetry of 2.51 fF with a far-corner artefact of -0.99 fF (Section VIII-E) shows that mesh noise can exceed the 0.3 fF claim. The ratio \(\varepsilon_{\mathrm{xtalk}}\) of (25b) is 0.9 \% at \(\Delta\)\(d\) = 0.1 mm, which is one LSB (0.1 fF) over an 11 fF signal, and 0.0014 and 0.0008 at 0.4 and 0.5 mm. Fig. 12 sets this against the open-field Far material difference of Section VIII-B (6.0--10.5 fF): grounding the ideal L2 sheet lowers the material term by more than an order of magnitude, down to the resolution of the simulation. Two further qualifications apply. First, the FEM screen is a continuous ideal equipotential L2 sheet (Section VIII); seams, discrete-pad gaps, the finite resistance of the L2 ground return and slot-to-slot switching transients are not represented, so a real 16-pad screen would increase \(c_{iT}^{\mathrm{res}}\) and the leakage budget for them remains open. Second, (25b) is electrostatic; the force-level acceptance (\(|\Delta F_{n}|\) \(\leq\) 0.1 N at 5 N) requires \(\mathcal{G}\) and the fixed-load proxy case has not been run.

\textbf{TABLE VI} Dual-Array Contact Mode: Metal (C1) and Nylon (C2) Probes, Whole-Layer L3 Compression

\begin{longtable}[]{@{}
  >{\raggedright\arraybackslash}p{(\columnwidth - 14\tabcolsep) * \real{0.1250}}
  >{\raggedright\arraybackslash}p{(\columnwidth - 14\tabcolsep) * \real{0.1250}}
  >{\raggedright\arraybackslash}p{(\columnwidth - 14\tabcolsep) * \real{0.1250}}
  >{\raggedright\arraybackslash}p{(\columnwidth - 14\tabcolsep) * \real{0.1250}}
  >{\raggedright\arraybackslash}p{(\columnwidth - 14\tabcolsep) * \real{0.1250}}
  >{\raggedright\arraybackslash}p{(\columnwidth - 14\tabcolsep) * \real{0.1250}}
  >{\raggedright\arraybackslash}p{(\columnwidth - 14\tabcolsep) * \real{0.1250}}
  >{\raggedright\arraybackslash}p{(\columnwidth - 14\tabcolsep) * \real{0.1250}}@{}}
\toprule\noalign{}
\begin{minipage}[b]{\linewidth}\raggedright
\(\Delta\)\(d\) (mm)
\end{minipage} & \begin{minipage}[b]{\linewidth}\raggedright
\(d_{0}-\Delta d\) (mm)
\end{minipage} & \begin{minipage}[b]{\linewidth}\raggedright
\(C_{\mathrm{metal}}\) (pF)
\end{minipage} & \begin{minipage}[b]{\linewidth}\raggedright
\(\Delta\)\(C\) (fF)
\end{minipage} & \begin{minipage}[b]{\linewidth}\raggedright
Interval slope (fF/mm)
\end{minipage} & \begin{minipage}[b]{\linewidth}\raggedright
\(C_{\mathrm{nylon}}\) (pF)
\end{minipage} & \begin{minipage}[b]{\linewidth}\raggedright
\(\Delta\)\(C_{\mathrm{leak}}\) (fF)
\end{minipage} & \begin{minipage}[b]{\linewidth}\raggedright
\(\varepsilon_{\mathrm{xtalk}}\)
\end{minipage} \\
\midrule\noalign{}
\endhead
\bottomrule\noalign{}
\endlastfoot
0 & 0.60 & 1.0769 & 0.0 & 92 & 1.0771 & -0.2 & --- \\
0.05 & 0.55 & 1.0815 & 4.6 & 128 & 1.0815 & 0.0 & 0 \\
0.1 & 0.50 & 1.0879 & 11.0 & 188 & 1.0878 & 0.1 & 0.0091 \\
0.2 & 0.40 & 1.1067 & 29.8 & 344 & 1.1067 & 0.0 & 0 \\
0.3 & 0.30 & 1.1411 & 64.2 & 738 & 1.1411 & 0.0 & 0 \\
0.4 & 0.20 & 1.2149 & 138.0 & 2320 & 1.2147 & 0.2 & 0.0014 \\
0.5 & 0.10 & 1.4469 & 370.0 & --- & 1.4466 & 0.3 & 0.0008 \\
\end{longtable}

\begin{figure}
\centering
\includegraphics{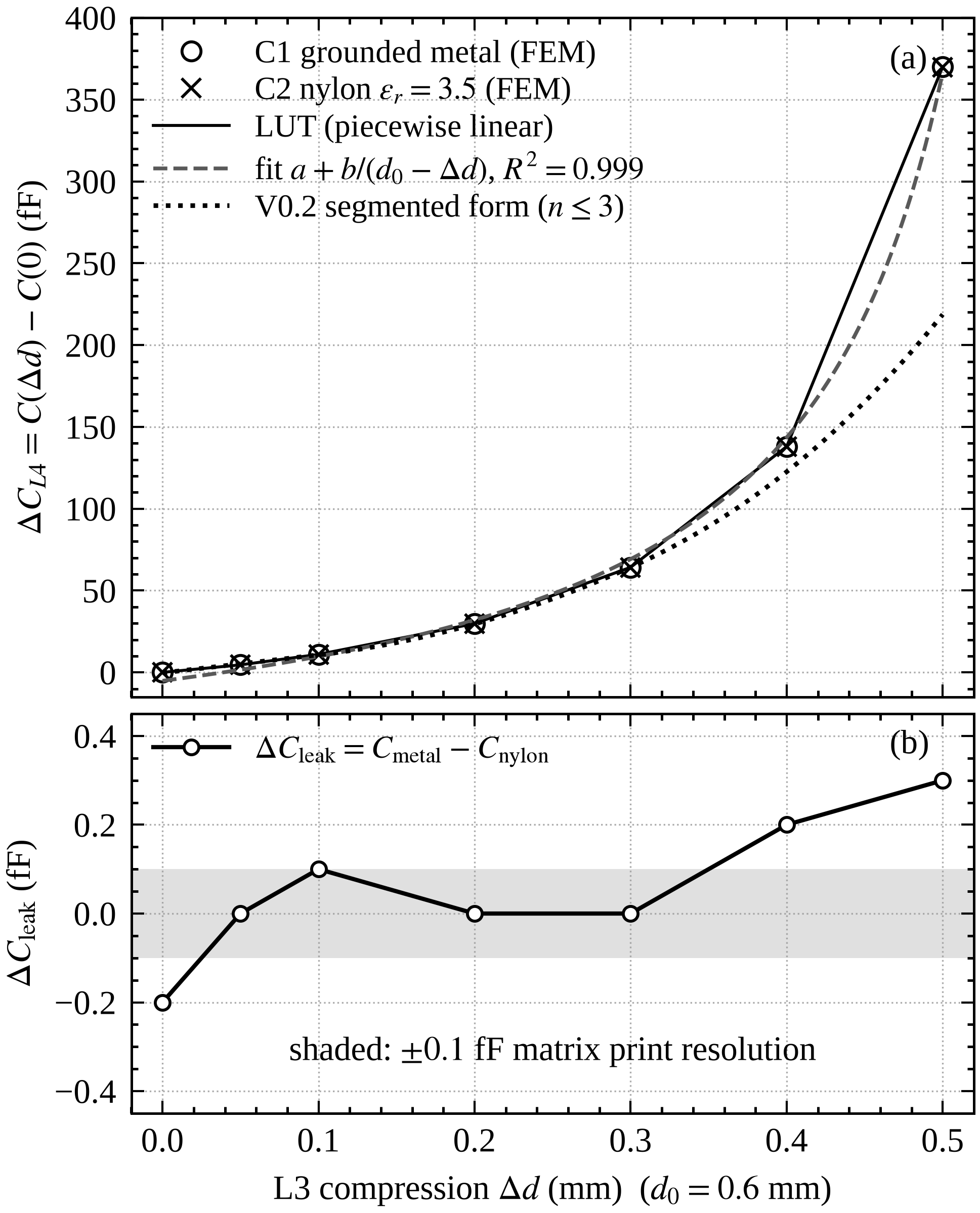}
\caption{Fig. 11}
\end{figure}

\emph{Fig. 11.} Dual-array Contact mode (FEM). (a) Compression response of L4{[}\(E_{11}\){]} for the metal (C1) and nylon (C2) probes, with the look-up-table interpolation, the single-term \(1/d\) fit and the segmented form inherited from the single-layer stack. (b) Material leakage \(\Delta\)\(C_{\mathrm{leak}}\); grey band: \(\pm\)0.1 fF print resolution.

\begin{figure}
\centering
\includegraphics{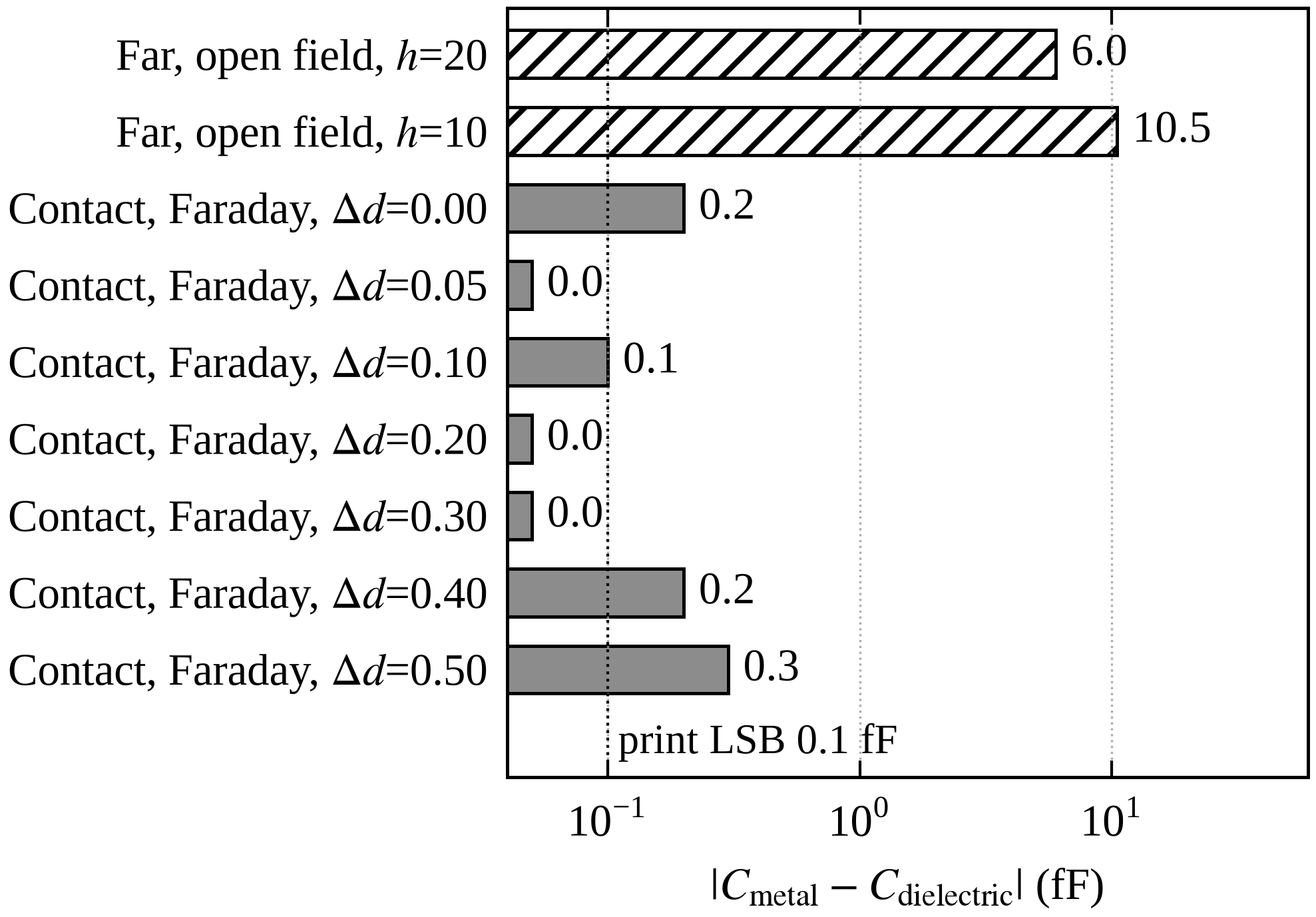}
\caption{Fig. 12}
\end{figure}

\emph{Fig. 12.} Magnitude of the metal-versus-dielectric reading difference in the open field (Far mode, plastic \(\varepsilon_{r}\) = 3.0 versus metal, F3) and behind the grounded L2 screen (Contact mode, nylon \(\varepsilon_{r}\) = 3.5 versus metal, C2). The two are different operators and are compared only in order of magnitude; the Far case is a proximity configuration, not a contact measurement.

\subsubsection{E. Dual-Array Contact: Neighbour Crosstalk (C4)}\label{e.-dual-array-contact-neighbour-crosstalk-c4}

C4 resolves all 16 L4 pads as separate 1 V ports with L2 grounded and the metal probe fixed 0.05 mm above the skin, and compresses only the 2 mm \(\times\) 2 mm cell of \(E_{11}\) (\(\Delta\)\(d\) = 0, 0.2, 0.4 mm; bottom rigid). Local and whole-layer compression give \(E_{11}\) readings within 1.8 fF of each other (\(\leq\) 0.15 \%), so the C1 curve represents a single-pixel press electrostatically. The mutual capacitances are resolved and symmetric (Table VII): the four edge neighbours couple with -2.366 to -2.380 fF at \(\Delta\)\(d\) = 0, falling to about -1.80 fF and -1.05 fF at 0.2 and 0.4 mm as the compressed \(E_{11}\) field terminates more on the grounded L2; the diagonal neighbours couple with about -0.16 fF and the eight farther pads with at most 0.0012 fF. If the neighbours were read as raw self-capacitances with the other pads at 0 V, the change of these terms would appear as crosstalk \(M_{\mathrm{mut}}=\Delta C_{\mathrm{mut},j}/\Delta C_{\mathrm{self},E_{11}}\) of 0.0185--0.0189 at 0.2 mm (Fig. 13a).

Under the circ readout all L4 pads are at the same potential and these mutual terms cancel identically, as (4) and (6a) require. What remains is the change of each neighbour's own capacitance to the grounded conductors. In this mesh it is not resolved: neighbour \(\Delta\)\(C_{\mathrm{circ}}\) values scatter between -0.99 and +2.07 fF at 0.2 mm (Fig. 13b), the far corner \(E_{03}\), which should be unaffected, reads -0.99 fF, and mirror-symmetric pairs differ by up to 2.51 fF. Averaging mirror pairs gives a maximum of 1.03 fF at 0.2 mm (3.1 \% of the 33.26 fF \(E_{11}\) change) and 1.58 fF at 0.4 mm (1.1 \% of 142.23 fF). The electrical crosstalk of the circ readout is therefore about 1 fF at the convergence level of the current coarse mesh. It was frozen at that figure without refinement in this revision because re-meshing the 16-port Contact model is computationally costly; the \textasciitilde1 fF value is consequently an observation under the present Track-A mesh, not a proven physical upper bound, and the same non-convergence caveat applies as for the 0.3 fF material-leakage observation. Mechanical crosstalk --- the compression of neighbouring L3 cells under a single-pixel load --- is expected to dominate but has no data.

\textbf{TABLE VII} Contact Crosstalk Terms for a Press on \(E_{11}\) (C4, 16-Port Model)

\begin{longtable}[]{@{}
  >{\raggedright\arraybackslash}p{(\columnwidth - 6\tabcolsep) * \real{0.2500}}
  >{\raggedright\arraybackslash}p{(\columnwidth - 6\tabcolsep) * \real{0.2500}}
  >{\raggedright\arraybackslash}p{(\columnwidth - 6\tabcolsep) * \real{0.2500}}
  >{\raggedright\arraybackslash}p{(\columnwidth - 6\tabcolsep) * \real{0.2500}}@{}}
\toprule\noalign{}
\begin{minipage}[b]{\linewidth}\raggedright
Quantity
\end{minipage} & \begin{minipage}[b]{\linewidth}\raggedright
\(\Delta\)\(d\) = 0
\end{minipage} & \begin{minipage}[b]{\linewidth}\raggedright
\(\Delta\)\(d\) = 0.2 mm
\end{minipage} & \begin{minipage}[b]{\linewidth}\raggedright
\(\Delta\)\(d\) = 0.4 mm
\end{minipage} \\
\midrule\noalign{}
\endhead
\bottomrule\noalign{}
\endlastfoot
\(E_{11}\) raw self \(C_{kk}\) (pF) & 1.0765514 & 1.1073425 & 1.2131121 \\
\(E_{11}\) minus C1 whole-layer reading (fF) & -0.35 & +0.64 & -1.79 \\
\(E_{11}\) \(\Delta\)\(C_{\mathrm{circ}}\) (fF) & --- & 33.26 & 142.23 \\
Edge neighbours \(C(E_{j},E_{11})\) (fF) & -2.366 \ldots{} -2.380 & -1.797 \ldots{} -1.802 & -1.044 \ldots{} -1.052 \\
Diagonal neighbours \(C(E_{j},E_{11})\) (fF) & -0.156 \ldots{} -0.158 & -0.114 \ldots{} -0.115 & -0.066 \ldots{} -0.067 \\
\(M_{\mathrm{mut}}\), edge (raw-self readout) & --- & 0.0185--0.0189 & 0.0096--0.0098 \\
\(M_{\mathrm{mut}}\), diagonal (raw-self readout) & --- & 0.0014 & 0.0007 \\
Circ crosstalk bound (mirror-pair mean, fF) & --- & 1.03 (3.1 \%) & 1.58 (1.1 \%) \\
\end{longtable}

\begin{figure}
\centering
\includegraphics{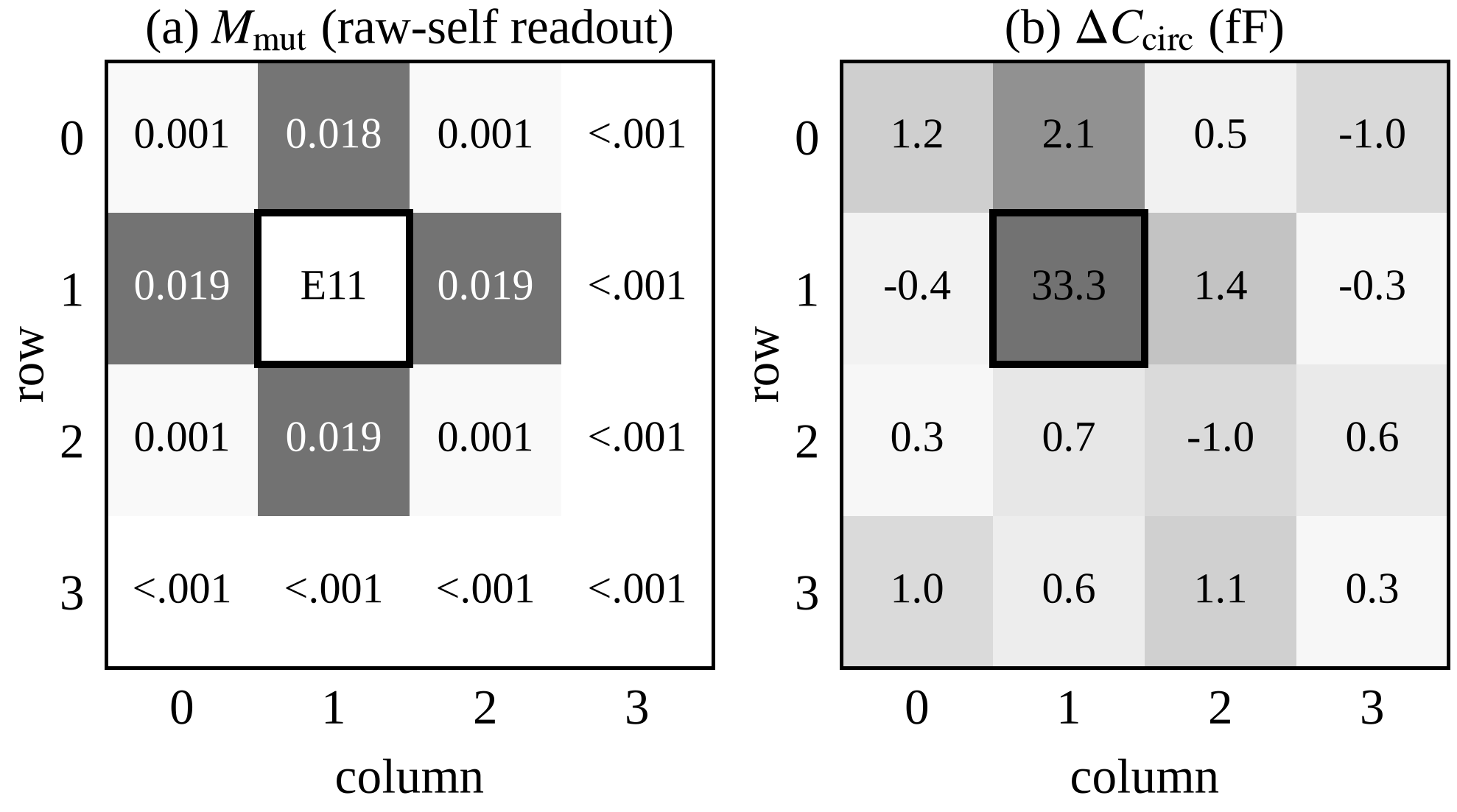}
\caption{Fig. 13}
\end{figure}

\emph{Fig. 13.} Neighbour response to a press of \(E_{11}\) (\(\Delta\)\(d\) = 0.2 mm, C4). (a) Mutual-capacitance crosstalk \(M_{\mathrm{mut}}\) that a raw-self readout would see. (b) \(\Delta\)\(C_{\mathrm{circ}}\) of every pad under the driven-shield readout; neighbour values are mesh-noise-limited (scatter -0.99 to +2.07 fF; mirror-pair mean \(\leq\) 1.03 fF). The bold frame marks \(E_{11}\).

\subsubsection{F. What the FEM Does Not Establish}\label{f.-what-the-fem-does-not-establish}

The dual-array campaign closes the electrostatic part of (6a)--(6c) and (25b) for the baseline geometry. It does not establish: the force map \(\mathcal{G}\) and \(F(\Delta d)\), hence no force resolution and no \(|\Delta F_{n}|\) figure; mechanical crosstalk; the Far multi-pad centroid; the orthogonal Mid pair and patch-wise compression variants, which were not run; the leakage of a real L2 screen with seams and finite ground impedance; the follower gain error \(\varepsilon_{A}\), \(\theta_{rate}\), the bench value of \(t_{set}\) and the frame rate; and the Far non-regression with respect to the single-layer lock. Mesh convergence is also incomplete (coarse mesh, non-convergence warnings). Each of these is listed as an open item in Table VIII.

\subsection{IX. Error Budget: Condensing the Claims Into Testable Items}\label{ix.-error-budget-condensing-the-claims-into-testable-items}

The noise variance decomposes as

\[\sigma_{C}^{2}=\sigma_{\mathrm{CDC}}^{2}+\sigma_{\mathrm{EMI}}^{2}+\sigma_{\mathrm{drift}}^{2}+\sigma_{\mathrm{sw}}^{2}.\tag{29}\]

\[\mathrm{SNR}=\frac{|\Delta C|}{\sigma_{C}},\qquad \delta F_{n}\approx\mathcal{G}'(u)\,\sigma_{C,\mathrm{sum}}.\tag{30}\]

The terms have distinct physical origins: \(\sigma_{CDC}\) comes from quantisation and excitation noise and is reduced by conversion time and averaging; \(\sigma_{EMI}\) comes from servos and mains interference and is reduced by choosing the excitation frequency away from harmonics, synchronous demodulation and shielding; \(\sigma_{drift}\) comes from temperature modulating \(\varepsilon_{r}\) and the geometry and is handled by temperature sensing and contact-free baseline tracking; \(\sigma_{sw}\) comes from MUX charge injection and is handled by reference capacitors and sufficient settling. The proximity threshold should be set at several times the unloaded \(\sigma_{C}\), and the force resolution maps the capacitance noise to newtons through \(\mathcal{G}'\). None of the four terms of (29) is available from electrostatic FEM; they must be measured.

The deterministic terms that FEM and the model of Section III do provide are collected in Table IX and Fig. 14. In Far mode the signals (3.1--15.9 fF) and the error terms (about 1--10 fF) overlap: the open-field material difference alone is of the same order as the 20 \(\rightarrow\) 10 mm increment, and a 10\(^{-3}\) follower mismatch costs half of it. The Far channel is therefore a decision channel, consistent with Section IV-B, and its distance output cannot be material-independent. In Contact mode the smallest simulated increment (4.6 fF at 0.05 mm) exceeds every electrostatic error term except the residual of a closed-form fit, which the look-up table removes; the contact budget is then set by the terms that FEM cannot supply --- \(\sigma_{C}\) of (29), mechanical crosstalk, the leakage of a real L2 screen and the uncertainty of \(\mathcal{G}\).

\textbf{TABLE IX} Deterministic Signal and Error Terms of the Dual-Array Stack (fF)

\begin{longtable}[]{@{}
  >{\raggedright\arraybackslash}p{(\columnwidth - 6\tabcolsep) * \real{0.2500}}
  >{\raggedright\arraybackslash}p{(\columnwidth - 6\tabcolsep) * \real{0.2500}}
  >{\raggedright\arraybackslash}p{(\columnwidth - 6\tabcolsep) * \real{0.2500}}
  >{\raggedright\arraybackslash}p{(\columnwidth - 6\tabcolsep) * \real{0.2500}}@{}}
\toprule\noalign{}
\begin{minipage}[b]{\linewidth}\raggedright
Mode
\end{minipage} & \begin{minipage}[b]{\linewidth}\raggedright
Term
\end{minipage} & \begin{minipage}[b]{\linewidth}\raggedright
Magnitude (fF)
\end{minipage} & \begin{minipage}[b]{\linewidth}\raggedright
Origin
\end{minipage} \\
\midrule\noalign{}
\endhead
\bottomrule\noalign{}
\endlastfoot
Far & Signal \(C\)(5) - \(C\)(20), raw & 15.9 & FEM F1 \\
Far & Signal \(C\)(10) - \(C\)(20), raw & 3.1 & FEM F1 \\
Far & Plastic vs metal at \(h\) = 10 mm & 10.5 & FEM F3 \\
Far & Mismatch leakage at \(\lvert\varepsilon_{A}\rvert\) = 10\(^{-3}\) & 1.47 & model (14), \(C_{eS}\) from FEM S0-F \\
Far & Offset-induced change at \(h\) = 20/10 mm & \textasciitilde1 & FEM F2 \\
Contact & Signal \(\Delta\)\(d\) = 0.05 mm & 4.6 & FEM C1 \\
Contact & Signal \(\Delta\)\(d\) = 0.2 mm & 29.8 & FEM C1 \\
Contact & Material leakage, max \(\lvert\Delta C_{\mathrm{leak}}\rvert\) & 0.3 & FEM C2 (print resolution; not mesh-converged) \\
Contact & Electrical crosstalk, circ readout & \textasciitilde1 (observation) & FEM C4 (current-mesh level; not refined) \\
Contact & Local vs whole-layer compression & 1.8 & FEM C4 vs C1 \\
Contact & Max residual of \(1/d\) closed form & 5.57 & FEM C1 fit (removed by LUT) \\
Contact & Mismatch leakage at \(\lvert\varepsilon_{A}\rvert\) = 10\(^{-3}\) & 0.01 & model (14), \(C_{eS}\) from FEM S0-C \\
Both & \(\sigma_{\mathrm{CDC}}\), \(\sigma_{\mathrm{EMI}}\), \(\sigma_{\mathrm{drift}}\), \(\sigma_{\mathrm{sw}}\) & not available & bench \\
Contact & Mechanical crosstalk; real-screen (seam) leakage & not available & mechanical FEM / bench \\
\end{longtable}

\begin{figure}
\centering
\includegraphics{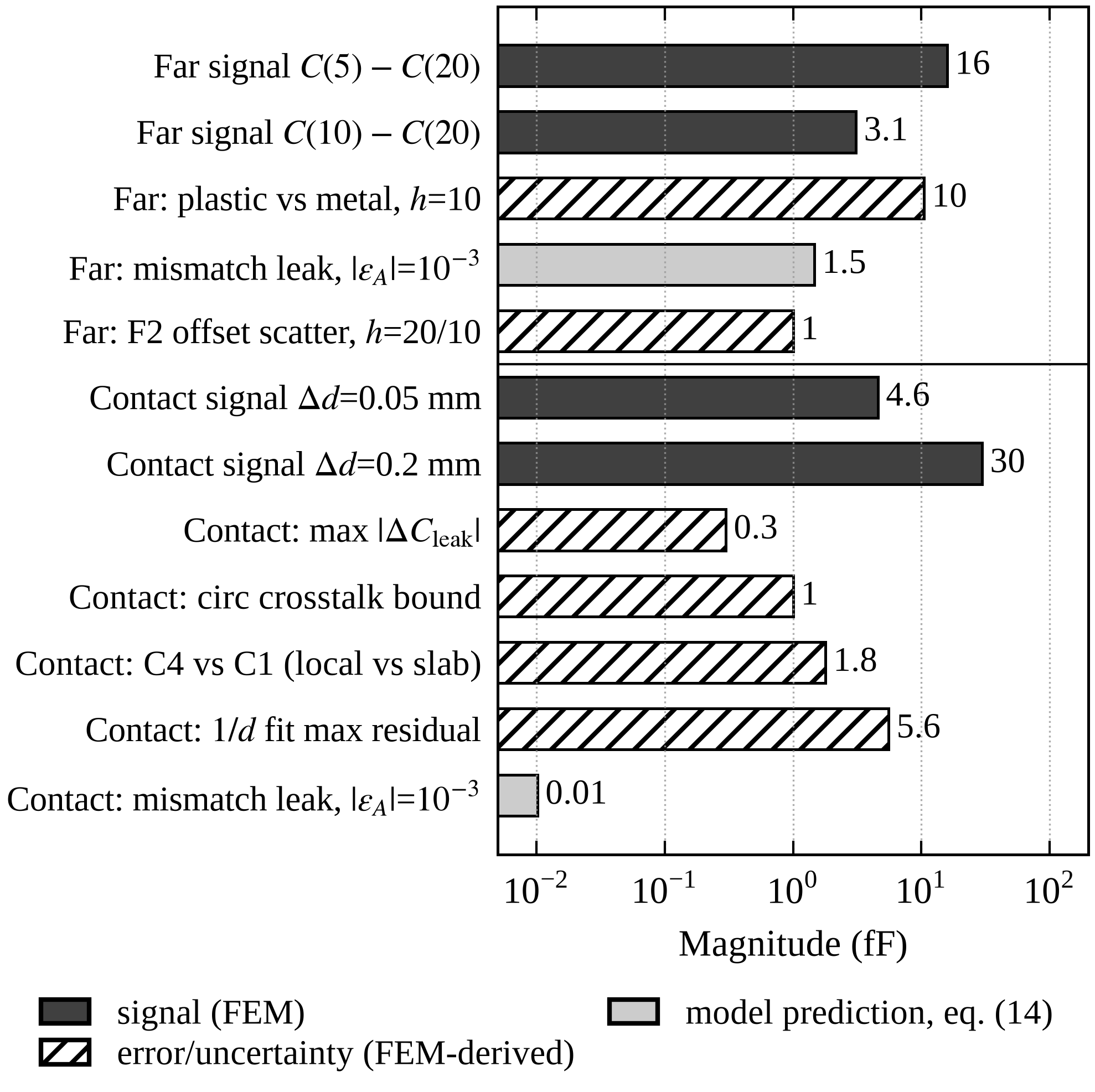}
\caption{Fig. 14}
\end{figure}

\emph{Fig. 14.} Deterministic signal and error terms of Table IX on a logarithmic scale. Dark: FEM signals; hatched: FEM-derived errors and uncertainties; light grey: model predictions of (14) with FEM-extracted \(C_{eS}\). Stochastic terms of (29) are not shown.

The testable propositions corresponding to (6)--(15), (16)--(18), (21)--(24) and (25)--(27) are condensed into Table VIII. The magnitudes in the table only indicate whether each term can become the dominant error; they are not pass/fail criteria. The last column records the status after the dual-array FEM campaign.

\textbf{TABLE VIII} Observation Equations and Testable Propositions

\begin{longtable}[]{@{}
  >{\raggedright\arraybackslash}p{(\columnwidth - 6\tabcolsep) * \real{0.2500}}
  >{\raggedright\arraybackslash}p{(\columnwidth - 6\tabcolsep) * \real{0.2500}}
  >{\raggedright\arraybackslash}p{(\columnwidth - 6\tabcolsep) * \real{0.2500}}
  >{\raggedright\arraybackslash}p{(\columnwidth - 6\tabcolsep) * \real{0.2500}}@{}}
\toprule\noalign{}
\begin{minipage}[b]{\linewidth}\raggedright
Claim
\end{minipage} & \begin{minipage}[b]{\linewidth}\raggedright
Governing equation(s)
\end{minipage} & \begin{minipage}[b]{\linewidth}\raggedright
Experimental implication
\end{minipage} & \begin{minipage}[b]{\linewidth}\raggedright
Status (this revision)
\end{minipage} \\
\midrule\noalign{}
\endhead
\bottomrule\noalign{}
\endlastfoot
Bone-mounted proximity is drowned by parasitics without shielding & (7), (8) & Shield on/off comparison in the same geometry & FEM (single-layer): 0.21 \% vs 6.12 \%; bench open \\
Mismatch leakage & (13)--(15) & Sweep follower gain; fit leakage slope & \(C_{eS}\) from FEM; \(\varepsilon_{A}\) not measured \\
Proximity increment monotonic in h; ill-posed far end & (16)--(18) & \(\Delta C\)(h) of a grounded sphere versus noise floor & FEM monotonic (\(R\) = 5 mm); noise floor open \\
Unbiased centroid requires footprint wider than pitch & (19) & Compare position error at centre vs.~edge & open (Far centroid not evaluated) \\
Crosstalk matrix depends on neighbour-electrode potential & (20) & Measure grounded scan and Vs-tied scan separately & FEM: mutual terms cancel in circ; \(\lesssim\) 1 fF at current-mesh level \\
Termination enters the operator; raw diagonals lose the differential & (6a), (6b) & Compare raw and circ extraction of the same channel & FEM Mid: raw fails, circ recovers at \(h\) \(\leq\) 3 mm \\
Insufficient settling appears as inter-slot crosstalk & (21)--(24) & Shorten settling time and watch threshold jitter & model; locked 3--5 \(\mu\)s; bench open \\
Contact force nonlinear; linear form not extrapolable & (25)--(27) & Load--unload curve and polynomial residuals & FEM \(C(\Delta d)\) nonlinear (25\(\times\)); \(\mathcal{G}\) open \\
Faraday screen removes target-material dependence of contact reading & (6c), (25a), (25b) & Nylon vs metal probe at the same 5 N load and position & FEM: \(\leq\) 0.3 fF at current mesh/print resolution (ideal continuous screen; not mesh-converged); \(\Delta F_{n}\) open \\
Centroid azimuth \(\theta\) is not tangential force & (28) & Shear comparison under varying preload & open \\
\end{longtable}

For fingertip geometry the expected capacitance magnitudes are: an unloaded single-pixel reading with shield \(C_{meas,0}\) \(\approx\) 1--4 pF; \(C_{SG}\) \(\approx\) 5--20 pF; unshielded \(c_{iG}\) \(\approx\) 10--40 pF; the paralleled \(\Delta C_{obj}\) for a conductive target at about 15 mm of tens to a hundred-odd femtofarads; and the \(\Delta C\) caused by newton-level contact of about one hundred to several hundred femtofarads. The dual-array FEM supports the first two (1.0763 pF for a single L4 pixel in Contact mode, 2.6168 pF for the paralleled L2 array in Far mode; shield-node self-capacitance 16.0--17.1 pF). The unshielded single-layer control (2.2 pF for the paralleled array) lies below the \(c_{iG}\) estimate. For the \(R\) = 5 mm probe the Far increment at 15 mm is only 1.7 fF, so the proximity estimate holds, if at all, only for finger- or hand-sized targets and is unvalidated. The contact range of 4.6--370 fF spans compressions of 0.05--0.5 mm; which forces these correspond to depends on \(\mathcal{G}\). These numbers are used to judge whether (15) and the thresholds can hold at all, and must be replaced by measurements of the actual stack-up and CDC.

\subsection{X. Discussion}\label{x.-discussion}

\subsubsection{A. Relation to Prior Work}\label{a.-relation-to-prior-work}

Equipotential shielding descends from the Capaciflector {[}1{]}, {[}2{]}; the partial-capacitance and image/finite-element cross-validation descends from proximity-fuze analysis {[}3{]}--{[}5{]} (related capacitive proximity-fuze control methods are also documented in {[}23{]}). On the robotics side, capacitive skins long focused on contact pressure {[}6{]}, {[}7{]}; proximity or dual-mode proximity--tactile sensing has developed through electrode reconfiguration {[}8{]}, arm/gripper deployment {[}9{]}--{[}12{]}, shear--proximity integration in soft structures {[}13{]}, fringing-field dual-mode sensors {[}14{]} and three-mode arrays with switchable shields {[}15{]}. Fingertip-scale multi-axis force and distributed capacitive tactile sensing continue to advance {[}16{]}, {[}17{]}. This paper claims neither the invention of the driven shield nor that of dual-mode capacitance; it places these elements under the constraint ``metal phalanx + thin silicone skin + shared CDC'', writes the CDC reading as an explicit decomposition, and points out that mismatch, boundary conditions and settling time are three independent error sources. Grounded screening layers are likewise an established means of rejecting external fields and target-dielectric influence in capacitive tactile sensors {[}24{]}, {[}25{]}; what is analysed here is the use of one electrode layer alternately as an outward-looking sensor and as the grounded screen, switched inside the same TDM frame, and the quantitative consequence of that switch for material cross-sensitivity.

\textbf{Dual-layer self-capacitance switching (Fukuoka University / Tsuji).} JP2016167252A (Example 5) and the accompanying papers {[}18{]}, {[}12{]} disclose a dual-layer self-capacitance proximity--contact scheme in which the upper electrode switches between ground and the detection circuit, and the lower electrode switches between the detection circuit and an in-phase shield drive: during contact the upper electrode is grounded so that the compression reading is independent of the object's electrical properties; during proximity the lower electrode acts as a driven shield. Later work combines self-capacitance proximity--tactile sensing with ToF and extends it to wrap-around string-like modules {[}26{]}, {[}27{]}. At the structural level this line already covers the core of ``outer grounded for force / inner in-phase shield for proximity''. The distinction of this paper is not to re-propose that topology, but to: (i) write the CDC reading as a Maxwell observation operator that includes the termination of every inactive electrode (eqs. (6a)--(6c)), with explicit cancellation of mutual terms under circ readout; (ii) give a quantitative mismatch-leakage gate (eqs. (14)--(15)) closed with FEM-extracted \(C_{eS}\) in a phalanx-dominated geometry; (iii) cast TDM settling as a frame-rate constraint (Section VI) rather than a mere switching narrative; and (iv) separate Far / Mid / Contact operators in one framework and quantify Faraday-stack suppression of material cross-sensitivity (Table VI, Fig. 12). In short, the Tsuji line provides a working dual-layer switch; this paper provides a testable budget of ``what is measured and where it errs'' under a bone-mounted constraint.

\textbf{Outer-electrode grounding on bionic fingertips (Tashan / Fogale / Cypress).} WO2025103388A1 {[}19{]} discloses a bionic mechanical finger in which the upper electrode senses proximity by self- or mutual capacitance before contact and is grounded as a shield after contact, while the lower electrodes are time-multiplexed for self-capacitance force sensing; an active shield may be placed on the PCB bottom. Fogale US10976460B2 {[}20{]} places a measurement electrode and a guard electrode across an elastic dielectric, applies a common alternating potential for approach/contact sensing, and an inter-electrode potential difference for pressure; unselected electrodes may be held at the active-electrode potential. Cypress US9490804B2 {[}28{]} grounds a conductive touch surface via a MUX once a proximity threshold is crossed, and reads compression on the electrode array beneath. All three support the structural proposition ``outer grounded after contact, inner reads compression''. Relative to them, this paper emphasises: (a) driven-shielding of the entire inner array in Far mode, with outer electrodes paralleled for SNR; (b) guard termination of inactive inner electrodes during contact scanning so that the crosstalk matrix enters the observation model; and (c) FEM evidence that, at the print resolution of the current non-converged mesh and under an ideal continuous L2 screen, the metal-versus-nylon difference falls to \(\leq\) 0.3 fF --- an observation, not a mesh-converged physical bound --- with discrete-pad seams and finite ground impedance left for the bench. In Fogale's pressure step the guard (inner layer) is grounded --- the opposite role assignment to the Faraday outer screen analysed here --- and the measurand differs accordingly.

\textbf{Reconfigurable multimodal skins and three-mode arrays.} CN118654713B {[}29{]} switches among ToF, capacitive proximity and contact modalities by distance thresholds and supports far merging / near splitting of electrodes; the far range relies on ToF. CN114322736A {[}30{]} switches a shared electrode array between distance sensing (electric-field Tx/Rx) and pressure sensing (array-to-ground-electrode CDC). Reference {[}15{]} trades detection depth against resolution with an iris-like switchable shield. This paper introduces neither ToF nor field shaping as its primary goal; it asks how small the shield mismatch must be, and how complete the TDM settling must be, when bone-mounted parasitics exceed the useful signal by one to three orders of magnitude. It is complementary to {[}15{]}: one answers ``how the field is actively shaped'', the other ``what the CDC actually measures after shaping''.

Electric-field pretouch and differential centering appear in Mayton et al.~{[}31{]}: the difference of two finger readings drives arm centering, with long-/mid-range servo switching on thresholds. The Mid operator of this paper forms a hardware differential on the outer array of one fingertip while the inner array remains a driven shield; that differs from software subtraction across separate finger electrodes and is not written here as a product state machine.

The difference from iCub-class contact capacitive fingertips {[}6{]}, {[}7{]} is the observation regime. Those systems treat a conductive cover layer as the other plate of the capacitor and optimise pressure hysteresis and covering durability; proximity is not their observation target. Only after replacing the cover by an insulating skin and introducing a driven shield can the same electrodes see targets outside the body again --- while re-introducing the parasitic problem of (7) and, if the same layer also measures force, the material cross-sensitivity of (25a). The dual-array stack can be read as restoring an iCub-like grounded cover during contact while keeping an insulating, outward-looking layer for proximity.

\textbf{Positioning summary.} Prior art already provides driven shields, dual-layer switching, outer-grounded Faraday force sensing, electrode merge/split and TDM scanning separately. The contribution of this paper concentrates on a unified observation operator under a metal-phalanx constraint, a quantitative shield-mismatch gate, FEM evidence, at the present mesh/print resolution and under an ideal continuous L2 screen, that a Faraday dual array reduces material cross-sensitivity, and a TDM error budget with a locked settling window --- a verifiable baseline for prototype calibration, not a product performance claim.

\subsubsection{B. Questions Not Answered Here}\label{b.-questions-not-answered-here}

The analysis does not replace experiments. Equation (16) is too coarse for a curved finger pad, a skin of finite thickness and the field distortion after paralleling; the FEM of Section VIII covers a planar baseline with a single probe size. The \(\varepsilon_{A}\) of (15) must be measured under the real \(C_{SG}\) load, not substituted by the follower's unloaded bandwidth. The force map \(\mathcal{G}\), mechanical crosstalk, the leakage of a real L2 screen with seams and finite ground impedance, and the frame rate are open. Tangential force, non-conductive targets in proximity mode, oil contamination and conductive dust are all outside the guarantee of this model. Turning a mechanistic analysis directly into product figures for proximity distance, force accuracy or frame rate mistakes an order-of-magnitude estimate for an acceptance criterion.

\subsubsection{C. Implications for Experiment Design}\label{c.-implications-for-experiment-design}

A minimal verification set can be read directly from Table VIII: shield on/off comparison; \(\varepsilon_{A}\) sweep with a linear fit of the leakage slope, performed in Far mode where \(C_{eS}\) is large; approach curve of a grounded sphere; centre/edge centroid comparison; \(M\) measured under both neighbour-electrode potentials and with raw versus circ extraction; settling-time sweep around the locked 3--5 \(\mu\)s window; force loading--unloading. The dual-array stack adds four items: the nylon-versus-metal probe test at the same 5 N load and position; a deliberate L2 seam or ground-impedance perturbation to measure the real screen leakage; single-pixel loading with neighbour readout to measure mechanical crosstalk; and a Far non-regression comparison against the single-layer lock under the same target. Missing any one of them leaves the corresponding claim at the order-of-magnitude level of this paper.

\subsection{XI. Conclusion}\label{xi.-conclusion}

For a capacitive fingertip array mounted on a metal phalanx, the CDC reading is a linear combination of the Maxwell capacitance matrix under given potential boundaries, not the electrode--target capacitance itself. Writing the termination of every inactive electrode into the observation operator makes this explicit: under the driven-shield (circ) readout the reading is the direct capacitance to all grounded conductors, mutual terms between equipotential electrodes cancel, and a channel extracted from raw diagonal entries can lose its signal --- as the dual-array Mid FEM shows. A driven shield forces \(V_{s}\) = \(V_{e}\), making the shield-branch current invisible to the measurement; with the FEM-extracted electrode--shield coupling of 1.47 pF in proximity mode, a mismatch of 10\(^{-3}\) already leaks half of the 3.1 fF increment from 20 mm to 10 mm, so the 10\(^{-3}\)--10\(^{-4}\) requirement is a hard Far-mode constraint, while the 10.2 fF coupling in contact mode makes mismatch negligible there. A single electrode layer that also measures force by skin compression reads the target's permittivity together with the force; a dual-array stack whose outer array is grounded during contact reduces this term to within 0.3 fF in FEM at the present mesh/print resolution (ideal continuous screen; not mesh-converged), while the inner array reads a monotonic, strongly nonlinear compression response (4.6--370 fF over 0.05--0.5 mm) that calls for a look-up table rather than a parallel-plate formula. TDM turns the follower load into settling time; at fingertip scale the first-order term is nanoseconds, and the locked 3--5 \(\mu\)s window is a margin for charge injection and follower dynamics that must be confirmed on the bench. The centroid azimuth cannot serve as a tangential force. The equations and FEM evidence above give testable constraints, not closed performance claims. Follow-up work should focus on the force map, mechanical crosstalk, real-screen leakage, mismatch sweeps and Far non-regression, rather than on extending a list of model variants.

\subsection{References}\label{references}

{[}1{]} J. M. Vranish and R. L. McConnell, ``Driven shielding capacitive proximity sensor,'' U.S. Patent 5 166 679, Nov.~24, 1992.

{[}2{]} J. M. Vranish, ``3-D capaciflector,'' U.S. Patent 5 726 581, Mar.~10, 1998.

{[}3{]} S. Cheng and Z. Cui, ``Theoretical analysis and numerical calculation of capacitive detector,'' J. Transducer Technol., vol.~23, no. 11, pp.~64--66, 2004 (in Chinese).

{[}4{]} S. Cheng, Z. Cui, and Z. Li, ``Theoretical analysis and numerical calculation of the electrode capacitance of a capacitive proximity fuze,'' J. Detection Control, vol.~25, no. S1, pp.~16--18, 32, 2003 (in Chinese).

{[}5{]} S. Cheng, Z. Cui, and X. Zhang, ``FE method of analysing detecting electrode of capacitance proximity fuze,'' J. Beijing Inst. Technol. (Engl. Ed.), vol.~11, no. 3, pp.~302--305, 2002.

{[}6{]} A. Schmitz, M. Maggiali, L. Natale, B. Bonino, and G. Metta, ``A tactile sensor for the fingertips of the humanoid robot iCub,'' in Proc. IEEE/RSJ Int. Conf. Intell. Robots Syst. (IROS), Taipei, Taiwan, 2010, pp.~2212--2217, doi: 10.1109/IROS.2010.5648838.

{[}7{]} A. Schmitz, P. Maiolino, M. Maggiali, L. Natale, G. Cannata, and G. Metta, ``Methods and technologies for the implementation of large-scale robot tactile sensors,'' IEEE Trans. Robot., vol.~27, no. 3, pp.~389--400, Jun.~2011, doi: 10.1109/TRO.2011.2132930.

{[}8{]} H. K. Lee, S. I. Chang, and E. Yoon, ``Dual-mode capacitive proximity sensor for robot application: Implementation of tactile and proximity sensing capability on a single polymer platform using shared electrodes,'' IEEE Sensors J., vol.~9, no. 12, pp.~1748--1755, Dec.~2009, doi: 10.1109/JSEN.2009.2030660.

{[}9{]} S. E. Navarro et al., ``Proximity perception in human-centered robotics: A survey on sensing systems and applications,'' IEEE Trans. Robot., vol.~38, no. 3, pp.~1599--1620, Jun.~2022, doi: 10.1109/TRO.2021.3111786.

{[}10{]} H. Wang, Z. Wang, G. Zhu, and S. Zhang, ``Capacitive proximity skin for non-contact liquid state measurement of robot grippers with different structures,'' Sens. Actuators A, Phys., vol.~365, 2024, Art. no. 114857.

{[}11{]} J. Sturm, M. Moradian, W. Scherr, A. Roshanghias, and H. Zangl, ``Integrated circuit for capacitive sensing in collaborative robotics,'' e\&i Elektrotech. Informationstech., vol.~142, pp.~377--387, 2025, doi: 10.1007/s00502-025-01341-1.

{[}12{]} S. Tsuji and T. Kohama, ``Sensor module combining time-of-flight with self-capacitance proximity and tactile sensors for robot,'' IEEE Sensors J., vol.~22, no. 1, pp.~858--866, Jan.~2022, doi: 10.1109/JSEN.2021.3130230.

{[}13{]} M. S. Sarwar et al., ``Touch, press and stroke: A soft capacitive sensor skin,'' Sci. Rep., vol.~13, 2023, Art. no. 17390, doi: 10.1038/s41598-023-43714-6.

{[}14{]} Y. Qin, H. Xu, S. Li, D. Xu, W. Zheng, W. Wang, and L. Gao, ``Dual-mode flexible capacitive sensor for proximity-tactile interface and wireless perception,'' IEEE Sensors J., vol.~22, no. 11, pp.~10446--10453, Jun.~2022, doi: 10.1109/JSEN.2022.3171218.

{[}15{]} X. Wu, Y. Xie, Z. Wu, et al., ``A bio-inspired proximity-tactile sensor array with beyond-extreme detection depth for embodied intelligence,'' Int. J. Extreme Manuf., vol.~8, no. 3, 2026, Art. no. 035506, doi: 10.1088/2631-7990/ae3ee6.

{[}16{]} Y. Song, J. Wang, et al., ``Fingertip-scale six-axis tactile interface with high-precision force sensing and position localization for dexterous human--machine interactions,'' Microsyst. Nanoeng., vol.~12, 2026, Art. no. 193, doi: 10.1038/s41378-026-01292-3.

{[}17{]} S. Ergun, R. B. Mishra, N. Anandan, T. Mitterer, V. Mattoli, and H. Zangl, ``CapTac: Robust capacitive sensing for distributed force mapping in parallel robotic grasping,'' IEEE Robot. Autom. Lett., vol.~11, no. 3, pp.~3668--3675, Mar.~2026.

{[}18{]} S. Tsuji and T. Kohama, ``Proximity and tactile sensor using self-capacitance measurement for human collaboration robots,'' IEEJ Trans. Sensors Micromachines, vol.~138, no. 1, pp.~2--8, Jan.~2018; see also JP2016167252A (Fukuoka Univ.), pub. Sep.~15, 2016.

{[}19{]} Beijing Tashan Technology Co., Ltd., ``Mechanical hand and bionic mechanical finger thereof,'' PCT Pub. WO2025103388A1, May 22, 2025; priority CN221291302U / CN221291301U, Jul.~9, 2024.

{[}20{]} C. Neel, F. Ossart, and D. Roziere, ``Device and method for detecting the approach and/or contact, and the pressure of an object in relation to a detection surface,'' U.S. Patent 10 976 460, Apr.~13, 2021.

{[}21{]} A. Slepyan, J. Chen, and N. Thakor, ``Scalable tactile sensing skins: Wiring and data management,'' Proc. IEEE, vol.~113, no. 11, pp.~1251--1269, Nov.~2025 (issue date; IEEE Xplore Date of Publication 12 Feb.~2026), doi: 10.1109/JPROC.2026.3660184.

{[}22{]} Texas Instruments, ``FDC1004 4-channel capacitance-to-digital converter,'' Data sheet SNOSCY5C, rev. C, Oct.~2024.

{[}23{]} S. Cheng et al., ``Capacitive proximity-fuze detection and control method for ground-skimming detonation,'' China Patent CN114594519B, Jun.~24, 2025 (in Chinese).

{[}24{]} B. Aksoy, Y. Hao, G. Grasso, K. M. Digumarti, V. Cacucciolo, and H. Shea, ``Shielded soft force sensors,'' Nat. Commun., vol.~13, 2022, Art. no. 4649, doi: 10.1038/s41467-022-32391-0.

{[}25{]} Pressure Profile Systems, ``Capacitive tactile sensor with nested matrix electrodes,'' U.S. Patent 9 983 744, May 29, 2018.

{[}26{]} S. Tsuji, ``String-like self-capacitance-based proximity and tactile sensor that can be wrapped around robotic arms,'' IEEJ Trans. Electr. Electron. Eng., vol.~20, no. 2, pp.~310--317, 2025, doi: 10.1002/tee.24191.

{[}27{]} S. Tsuji, ``Wrap-around string-like ToF and self-capacitance combined sensor for robots,'' IEEJ Trans. Electr. Electron. Eng., vol.~20, no. 5, pp.~688--695, 2025.

{[}28{]} Cypress Semiconductor Corp., ``Capacitance sensing circuits, methods and systems having conductive touch surface,'' U.S. Patent 9 490 804, Nov.~8, 2016.

{[}29{]} Zhejiang Univ. and China Ordnance Industry Automation Research Institute, ``Multimodal reconfigurable interactive flexible electronic skin,'' China Patent CN118654713B, Dec.~2, 2025 (pub. as CN118654713A, Sep.~17, 2024).

{[}30{]} Beihang Univ. and Goertek (Weifang) Intelligent Robotics, ``Distance and pressure dual-mode capacitive sensing with a shared electrode array,'' China Patent CN114322736A, Apr.~12, 2022 (authorized B, Oct.~28, 2025).

{[}31{]} B. Mayton, L. LeGrand, and J. R. Smith, ``An electric field pretouch system for grasping and co-manipulation,'' in Proc. IEEE Int. Conf. Robot. Autom. (ICRA), Anchorage, AK, USA, 2010, pp.~831--838.

\end{document}